\documentclass[table,xcdraw]{article}
\usepackage[utf8]{inputenc}
\usepackage{graphicx} 
\usepackage[round, authoryear]{natbib}
\usepackage{geometry}
\usepackage{siunitx}
\usepackage{titlesec}
\usepackage{hyperref}
\titleformat{\paragraph}
{\normalfont\normalsize\bfseries}{\theparagraph}{1em}{}
\titlespacing*{\paragraph}{0pt}{3.25ex plus 1ex minus .2ex}{1.5ex plus .2ex}
\usepackage[parfill]{parskip}
\usepackage{commath}
\usepackage{multirow}
\usepackage{colortbl}
\usepackage{subcaption}
\usepackage{mathrsfs}
\usepackage{amssymb}

\usepackage{array}
\newcolumntype{L}{>{\centering\arraybackslash}m{3cm}}
\usepackage{amsmath}
\usepackage{mathtools}
\usepackage{lineno}
\usepackage{breqn}
\usepackage{bm}
\usepackage{tikz}
\usetikzlibrary{matrix,chains,positioning,decorations.pathreplacing,arrows}
\usepackage[makeroom]{cancel}
\usepackage{relsize}
\usepackage{changepage}
\usepackage{tcolorbox}
\usepackage{float}
\usepackage[labelfont={color=blue,bf}]{caption}
\usepackage{setspace}
\usepackage{soul}
\usetikzlibrary{positioning, fit, arrows.meta, shapes}

\allowdisplaybreaks
\title{Advancing Spatiotemporal Prediction using Artificial Intelligence: Extending the Framework of Geographically and Temporally Weighted Neural Network (GTWNN) for Differing Geographical and Temporal Contexts}
\author{Nicholas Robert Fisk$^{1}$, Matthew Ng Kok Ming$^{2}$, Zahratu Shabrina$^{1}$\\
\normalsize{$^{1}$Department of Geography, King's College London, UK}\\
\normalsize{$^{2}$City Futures Research Centre, University of New South Wales, Australia}\\
\normalsize{$^\ast$To whom correspondence should be addressed; E-mail:  zara.shabrina@kcl.ac.uk.}}
\date{}

\begin{document}
\begin{singlespace}
\maketitle

\begin{abstract}
This paper aims at improving predictive crime models by extending the mathematical framework of Artificial Neural Networks (ANNs) tailored to general spatiotemporal problems and appropriately applying them. Recent advancements in the geospatial-temporal modelling field have focused on the inclusion of geographical weighting in their deep learning models to account for nonspatial stationarity, which is often apparent in spatial data. We formulate a novel semi-analytical approach to solving Geographically and Temporally Weighted Regression (GTWR), and applying it to London crime data. The results produce high-accuracy predictive evaluation scores that affirm the validity of the assumptions and approximations in the approach. This paper presents mathematical advances to the Geographically and Temporally Weighted Neural Network (GTWNN) framework, which offers a novel contribution to the field.  Insights from past literature are harmoniously employed with the assumptions and approximations to generate three mathematical extensions to GTWNN's framework. Combinations of these extensions produce five novel ANNs, applied to the London and Detroit datasets. The results suggest that one of the extensions is redundant and is generally surpassed by another extension, which we term the \textit{history-dependent module}. The remaining extensions form three novel ANN designs that pose potential GTWNN improvements. We evaluated the efficacy of various models in both the London and Detroit crime datasets, highlighting the importance of accounting for specific geographic and temporal characteristics when selecting modelling strategies to improve model suitability. In general, the proposed methods provide the foundations for a more context-aware, accurate, and robust ANN approach in spatio-temporal modelling.

\end{abstract}
\maketitle

\section{Introduction}
The field of predictive crime modelling aims to further the tools afforded to law enforcement agencies for proactive resource allocation and directed patrols \citep{ferreira2012gis, rummens2017use}. The effectiveness of these crime intervention policies, when applied proactively, is highly dependent on the precision of the predictive models that inform the strategy. Various modelling approaches, ranging from simpler statistical models to sophisticated machine learning models, have been applied to the problem of crime prediction \citep{jenga2023machine, tamir2021crime, zhang2022interpretable}. Within the realm of policing strategies, various hypotheses have been proposed to understand the relationship between police interventions and crime prevention \citep{sherman2003policing}. These hypotheses encompass different orientations, each positing a distinct approach to reducing crime. Of particular interest to this study is the hypothesis of \textit{`Directed Patrols,'} which suggests that concentrating police efforts towards specific \textit{`hotspots'} at optimal times can significantly contribute to crime reduction leading to an enhanced the quality of life for urban residents, safer environments and community well-being, and increased confidence in law enforcement agencies.

Among the various machine learning approaches employed for crime prediction, Artificial Neural Networks (ANNs) have seen the most widespread use due to their versatility in customisation. With the recent widespread public success of OpenAI's ANN models: DALL-E (released Jan 2021, \citep{ramesh2022hierarchical}) and ChatGPT (released Nov 2022, \citep{radford2018improving}), the promotion of ANNs in scientific literature is expected to further increase. As crime prediction falls under the blanket of spatiotemporal problems, we invoke the state-of-the-art ANN model for general spatiotemporal problems, namely the Geographically and Temporally Weighted Neural Network (GTWNN) as introduced by \citep{wu2021geographically}. This paper aims to further develop the mathematical framework of GTWNN, and provide advances for the appropriate application of ANNs to crime prediction problems given intrinsic spatiotemporal statistical properties present in data. Consequently, the fundamental research question this paper aims to address is: \textit{How can we extend the mathematical framework of ANNs tailored to general spatio-temporal problems and appropriately apply them to crime prediction?}

This research question centres around extending the framework of GTWNN, and developing methodologies which aim to either increase the affordability of applying ANNs or inform a more appropriate prescription for their application. GTWNN, developed in 2021 by \cite{feng2021geographically} is an extension to the mathematical spatiotemporal model, Geographically and Temporally Weighted Regression (GTWR) as introduced by \cite{fotheringham2015geographical}). GTWR linearly combines contextual external factor information and is solved with the aid of kernel functions \citep{fotheringham2015geographical}. By incorporating ANNs in the model, GTWNN has two advantages over its parent equation: upgrading the linear combination of external factors to a non-linear combination and allowing the dependent variable to be modelled more generally without relying on the functional structure of kernel functions \citep{feng2021geographically}. We hypothesise two limitations of the GTWNN model: a lack of continuity along the spatial and temporal axes for the coefficient functions, referred to in the relevant literature as \textit{influence factors}, attached to external factors (which are suggested to be continuous in GTWR), and a lack of historical contextual information which can aid in the production of more accurate influence factors.


Beyond the appropriate concern for the meaningful incorporation of contextual theory, there exists an ever-increasing desire for higher quality neural networks with increased accuracy. Increased ANN accuracy can be achieved by: general improvements to the ANN's architectural design, improved data collection, an increased volume of real data, higher resolution data, augmenting and balancing data, incorporation of real statistically observable trends, and prescriptive schemes for their appropriate application. As a result, this paper aims to consider all these approaches except improved data collection and increased volume of real data, which are beyond the scope of this paper. Although this paper aims to increase the predictive capabilities of ANNs for crime prediction, there also exists considerable desires from other disciplines for increased predictive accuracy. The methodologies presented in this paper are not general to all types of prediction problems, however, they are generally applicable to prediction problems of a spatiotemporal nature.

\section{The Geography of Crime}
Crime prediction is often used to explain the geography of crime, referred as the analysis of the spatial distribution of crime to explain the patterns of crime \citep{fyfe2000geography}. Attempts towards explaining the distribution of crime have largely referenced theories regarding, human ecology, forms of urban management (e.g. policing), the housing market, and the general built environment. There exist many aspects by which crime Geographers choose to analyse crime. Some examples include: crimes against women with reference to their social segregation, and highlighting hidden crimes in domestic and private enclosures \citep{valentine1989geography}; and focussing on wildlife crimes and policing in rural areas \citep{fyfe2011thin}. The geography of crime is viewed by most researchers in the area as a niche field of research \citep{yarwood2015geography}, as research in that category cuts across many disciplines including, cultural and social geography and criminology using appropriate statistical analyses to infer correlations and significance.

The geography of crime focusses on understanding the relationships between crime, space, and society through the critical analysis of victims and offenders, and the impact of crime on society. The theoretical basis underpinning these relationships is given by a set of opportunity theories: routine activity theory, rational choice theory, and crime pattern theory. Routine activity theory suggests that there exists ``interdependence between the structure of illegal activities and the organisation of everyday sustenance activities'', and that the likelihood of crime at any specific place and time is a ``function of the convergence of likely offenders and suitable targets in the absence of capable guardians'' \citep{cohen1979social}. Rational choice theory, developed in 1986 by Cornish and Clarke \citep{cornish2014reasoning}, posits a risk versus reward argument, whereby risk is the probability of being caught. Consequently, the theory suggests that by increasing the risk of offending and decreasing the potential rewards, then offenders should be further deterred from committing an offence. Crime pattern theory, developed by Brantingham and Brantingham \citep{brantingham2013crime}, examines how offenders and victims cross paths, and thus the distribution of crimes across urban space. Through travel paths and activity nodes, people establish their own activity spaces \citep{golledge1987analytical}, together with awareness spaces that offenders use to choose their targets \citep{bernasco2005residential}. Activity spaces exhibit a directional bias or preference \citep{frank2012criminal}, in which offences are generally committed within some direction from their homes, towards notable crime attractor locations (e.g. bars, and shopping districts). In 2014, Reid et al. showed that a distance decay function is expressed not only at the activity nodes but also around the travel paths connecting the nodes \citep{reid2014uncovering}.

\section{Model Designs \& Architectures}

\begin{figure}[h]
\centering
\includegraphics[width=0.55 \linewidth]{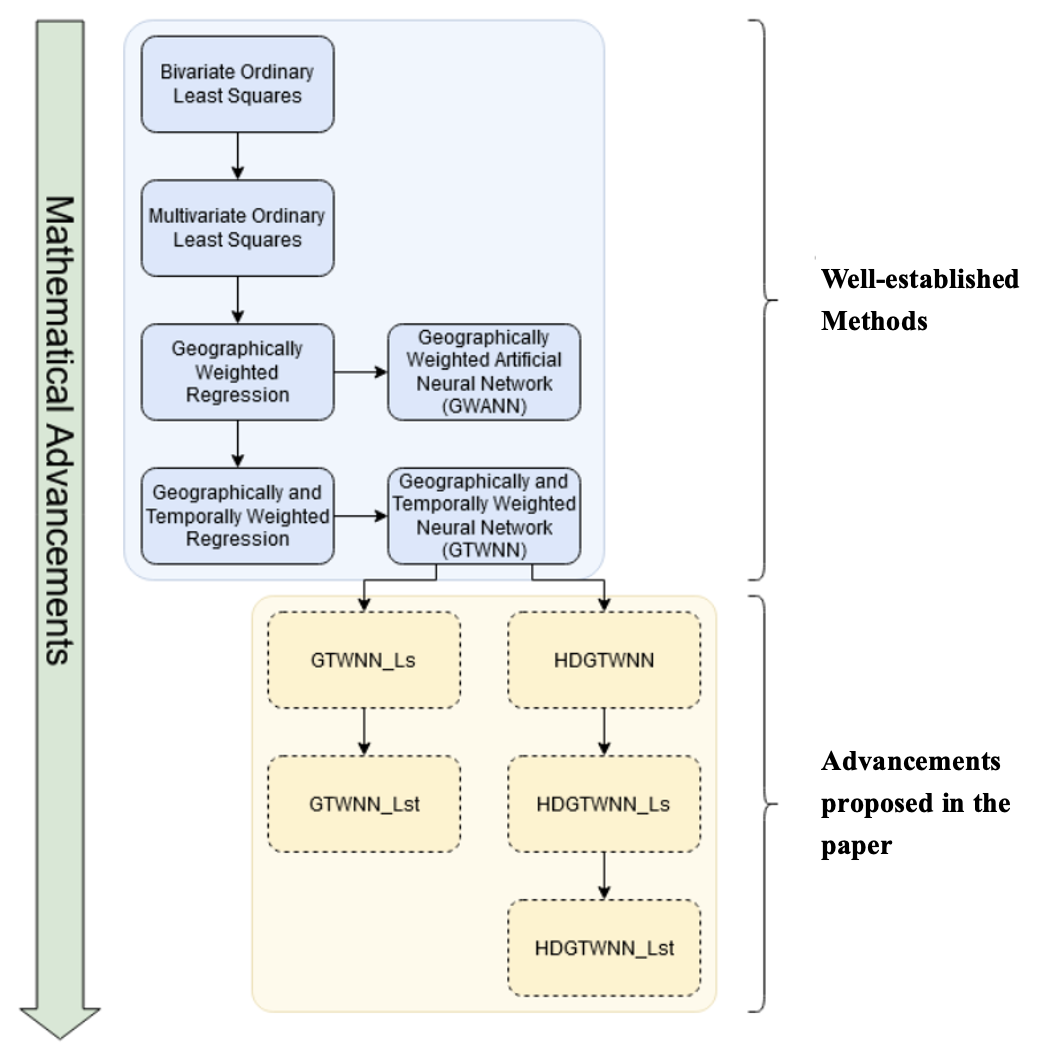} 
\end{figure}

\subsection{Vanilla NN}
To assess the potential shortcomings of attempting to relate a space-time coordinate ($x$=Longitude, $y$=Latitude, $t$=time-index) to a crime count $C$ for the coordinate we first applied a vanilla neural network. This type of network is characterised as a feed-forward fully connected dense neural network with several hidden layers, each comprising a varying number of neurons. The objective of predicting crime occurrences demands an ideal mapping, denoted as:
\begin{equation}
    f : (x,y,t) \xrightarrow{} C
\end{equation}
where $f$ is a function which perfectly maps each space-time coordinate to a crime count. The role of the vanilla neural network is to approximate this function through a computational approach, yielding an approximate mapping denoted as $\Tilde{f}$:
\begin{equation}
    \Tilde{f} : (x,y,t) \xrightarrow{} C + \varepsilon
\end{equation}
where $\varepsilon$ represents the error in the predicted crime count. Below is an illustrative representation depicting the application of the vanilla neural network to the crime prediction problem:
\begin{figure}[H]
  \centering
  \includegraphics[width=0.5\linewidth]{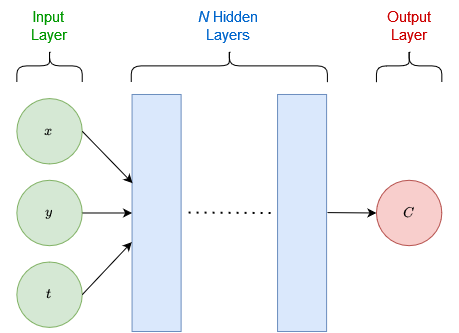}
  \caption{Pictorial representation of the vanilla NN model applied to crime prediction.}
\label{vanillaNNdiag}
\end{figure}

For the vanilla networks used in this section, we apply the the mean squared error (MSE) loss function.

\subsection{GWANN}

Hagenauer and Helbich introduced a novel artificial neural network, GWANN, inspired by the Geographically Weighted Regression (GWR) method \citep{fotheringham2015geographical}, as outlined in their research paper \citep{hagenauer2022geographically}. Two key design features/insights of their model are the primary focus of our investigation.

Firstly, they propose that since the influence factors of GWR are continuous in nature, the prediction for a cell at $(x,y,t)$ can offer valuable information about its surrounding cells, enabling the generation of approximate predictions for neighboring cells.

Secondly, they emphasise the significance of geographically weighted outputs and advocate explicitly incorporating this aspect into the loss function to induce the weighting effect during the training process. While the design of the output layer of GWANN and incorporation of the loss function differ between our model and theirs, the concept of a geographically weighted loss and the structural form of the loss function they present reflect similar broad goals between their network and our novel GWANN variant.  

We adopt our own adaptation of these two design features from the GWANN model to enhance the vanilla neural network presented in the preceding subsection. Below, we provide a visual representation of our GWANN model for reference:
\begin{figure}[H]
  \centering
  \includegraphics[width=0.7\linewidth]{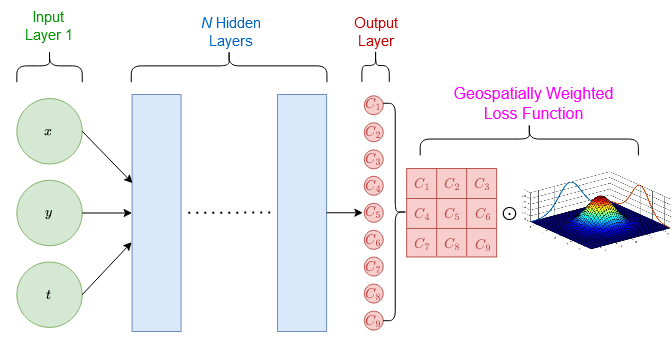}
  \caption{Pictorial representation of the GWANN model applied to crime prediction.}
\label{fig5-GWANN}
\end{figure}

For our variant of the GWANN networks used in this section we apply the loss function given by (\ref{lossfunc}).

For the output layer, we predict 9 values, which form a flattened array of a $3 \times 3$ grid, where the central cell corresponds to the estimated number of crimes for the input coordinate. The model is trained using the loss function $L$ defined as follows:
\begin{equation}
\label{lossfunc}
    L = \frac{1}{2} \sum_{i=1}^{n_i} v_i (t_i - o_i)^2
\end{equation}
In the above equation, $t_i$ and $o_i$ represent the target and model output, respectively. The summation is conducted over $n_i$ cells considered in the spatial vicinity of location $i$. The geographical weighting term $v_i$ is determined as:
\begin{equation}
    v_i = \exp\left\{-\frac{1}{2}\left(\frac{d_{ij}}{h}\right)^2\right\}
\end{equation}

It is important to note that our variant of GWANN deviates from the original approach. In paper by \cite{hagenauer2022geographically}, GWANN's output is not a prediction aimed at mapping coordinates to a future time point; rather, it associates coordinates with observations at the current time. The primary objective of this approach is to derive more accurate influence factors compared to those achievable through GWR and the methodologies described in \cite{brunsdon1998geographically}, while also adopting geographically weighted moving average for the target central cell more closely mimicking the way geographical weighting is applied in GWR. 

While we call our variant GWANN in this section of the thesis to maintain a consistent naming convention, it should be noted that the two networks present thematic differences. 
\subsection{GTWNN}

GTWNN \citep{feng2021geographically} currently stands, to our knowledge, as the state-of-the-art (SOTA) model for spatiotemporal prediction. It is built upon the foundation of the GTWR formula:
\begin{equation}
\label{GTWR}
    y = \beta_0 + \sum_{i=1}^{N} \beta_i x_i + \varepsilon
\end{equation}
where $y$ represents the dependent variable, ${x_1,\cdots,x_N}$ denotes the $N$ independent variables with their associated influence factors ${\beta_1,\cdots,\beta_N}$, $\beta_0$ captures any unaccounted information by the independent factors, and $\varepsilon$ denotes the error. Importantly, all these terms are spatiotemporal functions.

Their network structure is pictorially represented below:
\begin{figure}[H]
  \centering
  \includegraphics[width=0.5\linewidth]{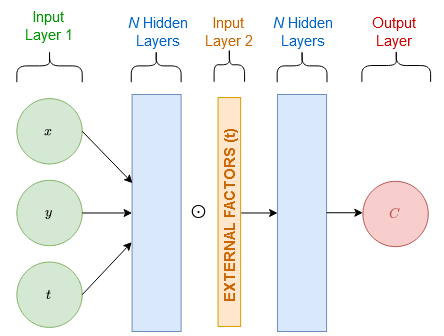}
  \caption{Pictorial representation of the GTWNN model applied to crime prediction. The circle-dot symbol represents element-wise multiplication.}
\label{fig5-GTWNN}
\end{figure}

The GTWNN model enhances the basic GTWR formula in the following aspects:
\begin{enumerate}
    \item The $\beta$ terms, which were conventionally determined using kernel functions (usually Gaussian), are no longer confined to the functional space of kernel functions.
    \item While GTWR entails a linear combination of the independent variables and their dependents, the second set of hidden layers in GTWNN elevates this to a non-linear combination of the independent variables.
\end{enumerate}
From the perspective of a practitioner, the introduction of these independent variables, also referred to as external factors, serves two additional purposes:
\begin{enumerate}
    \item The neural network gains access to more genuine information, which enriches its learning capability.
    \item The functional space of mappings that the neural network can establish between inputs and outputs is significantly reduced, as any formed relations are constrained to pass through the external factors.
\end{enumerate}

This network, named GTWNN, corresponds exactly to the GTWNN model presented in reference \citep{feng2021geographically}.

\subsection{GTWNN\_Ls}

Given that both GWANN and GTWNN models allude to positive extensions of the vanilla neural network, reducing validation errors through independent approaches, a natural progression is to merge the two models, creating a hybrid model. The pictorial representation of the hybrid model, applied to crime prediction, is shown below:
\begin{figure}[H]
  \centering
  \includegraphics[width=0.7\linewidth]{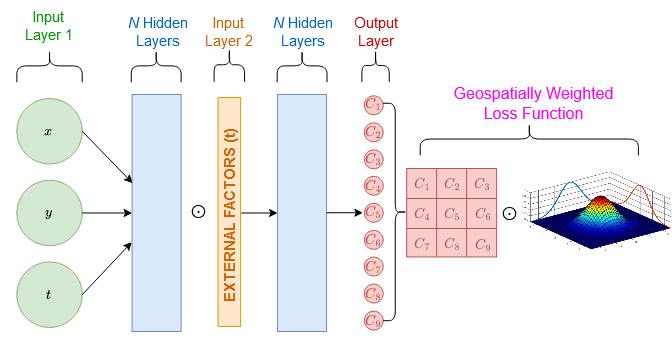}
  \caption{Pictorial representation of the hybrid model, GTWNN\_Ls, applied to crime prediction.}
\label{fig5-GTWNNN_Ls}
\end{figure}

In theory, this hybrid model is anticipated to offer the following advantage:

\begin{itemize}
    \item The output layer constrains the output of the first hidden layer block representing the set of $\beta$s to the space of continuous functions along the spatial axis. This is consistent with one of the primary assumptions of Geographically Weighted Regression (GWR), which posits that the functions representing the influence factors maintain continuity across different spatial locations. This is expected to result in enhanced generalisability to unseen test data.
\end{itemize}

\subsection{GTWNN\_Lst}

The proposed model, referred to as GTWNN\_Lst, shares similarities with GTWNN\_Ls but distinguishes itself through its output layer and loss function. A graphical representation of the GTWNN\_Lst architecture is presented below:

\begin{figure}[H]
  \centering
  \includegraphics[width=0.8\linewidth]{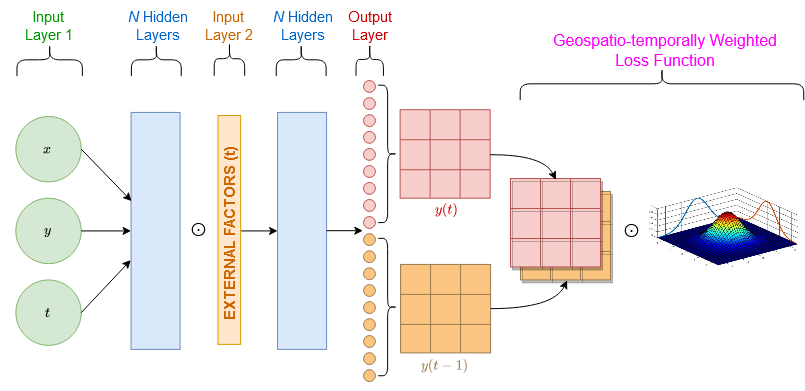}
  \caption{Pictorial representation of the GTWNN\_Lst model applied to crime prediction.}
\label{fig5-GTWNN_Lst}
\end{figure}

Given the spatiotemporal nature of our data and the assumption in GTWR that functions representing influence factors are continuous in both space and time, it is natural to seek a network that confines the set of $\beta$s to the domain of spatiotemporally continuous functions. Our objective is to enforce the network to generate $\beta$s that facilitate approximate predictions for spatio-temporally adjacent cells. This ensures that the influence factors produced by the network have spatio-temporal continuity.

\subsection{HDGTWNN}

The history dependent GTWNN model (HDGTWNN) exhibits considerable resemblance to GTWNN, albeit with a significant distinction, the incorporation of three input layers instead of two. The network's graphical depiction is presented below:

\begin{figure}[H]
  \centering
  \includegraphics[width=0.65\linewidth]{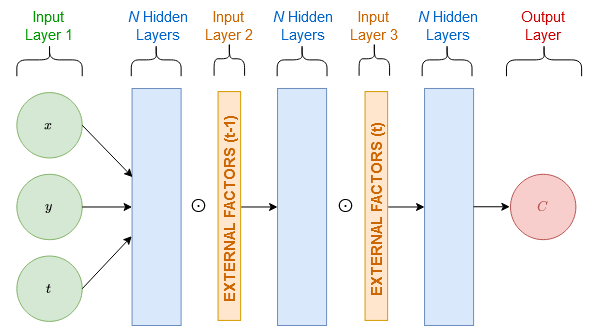}
  \caption{Pictorial representation of the history dependent GTWNN model (HDGTWNN) applied to crime prediction.}
\label{fig5-HDGTWNN}
\end{figure}

This model incorporates the history-dependent module, as detailed in Section 5.3.3, as an extension to the GTWNN architecture. By implementing this module, the network is expected to yield more accurate influence factor values through the generation of $\beta(t-1)$, which are combined with the external factors at time $(t-1)$ and subsequently evolved to time $t$ in the subsequent hidden layer block. In Section 5.3.4, we explore the model's ability to reliably evolve the $(t-1)$ product to time $t$ and propose that the second hidden layer block also functions to rectify any potential errors that may arise during the formation of $\beta(t-1)$ in the first hidden layer block.

\subsection{HDGTWNN\_Ls}

\begin{figure}[H]
  \centering
  \includegraphics[width=0.8\linewidth]{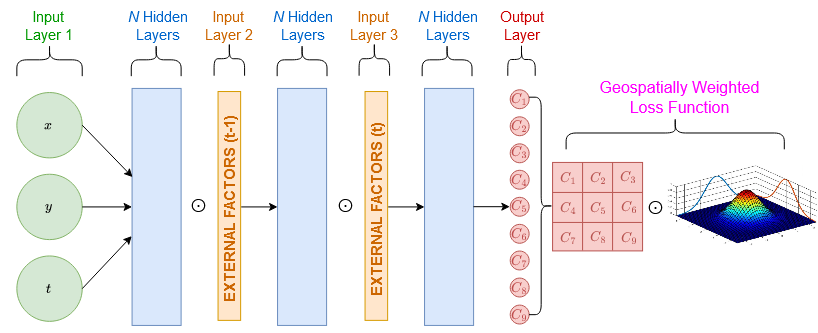}
  \caption{Pictorial representation of the HDGTWNN\_Ls model applied to crime prediction.}
\label{HDGTWNN_Ls}
\end{figure}

HDGTWNN\_Ls differs from HDGTWNN only in terms of its output layer and loss function. The visual representation of the model is presented in \textcolor{blue}{\textbf{Figure \ref{HDGTWNN_Ls}}}.

The advantages expected from HDGTWNN\_Ls are briefly discussed in sections 5.4.2 and 5.4.6, and explored in greater detail in sections 5.3.1 and 5.3.3. In essence, HDGTWNN\_Ls is projected to yield more precise $\beta$ values and confine them to the space of continuous functions along the spatial axis.

\subsection{HDGTWNN\_Lst}

HDGTWNN_Lst is an extension of HDGTWNN_Ls, which incorporates the additional features outlined in Section 5.3.2. A graphical representation of the HDGTWNN\_Lst model is presented below:

\begin{figure}[H]
  \centering
  \includegraphics[width=0.9\linewidth]{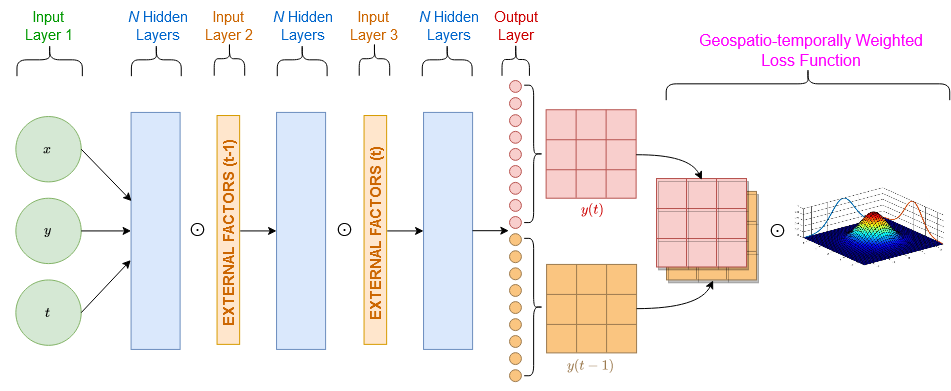}
  \caption{Pictorial representation of the HDGTWNN\_Lst model applied to crime prediction.}
\label{HDGTWNN_Lstdiag}
\end{figure}

The potential benefits of this network are discussed in sections 5.4.5 and 5.4.6, and explored in greater detail in sections 5.3.2 and 5.3.3. In essence, HDGTWNN\_Lst is expected to provide more accurate $\beta$ values, while constraining them to the space of continuous functions along both the spatial and temporal dimensions.

\section{Data}

\subsection{London Crime Data Pre-processing }

The London crime data set used in this study was collected from the archives of https://
data.police.uk/data/. The temporal span of the data covers 9 years, ranging from January 2011 to December 2019.  The temporal resolution of the data is set at monthly intervals, providing a monthly aggregation of crime records. Spatially, the dataset is represented in a longitude and latitude format, reflecting street-level crime occurrences.

Prior to preprocessing, the dataset initially contained a total of 9,451,027 entries. Subsequently, entries lacking essential information, such as spatial or temporal details, were removed, as were instances of mislabeled crime claims pertaining to non-London boroughs. After the elimination of erroneous entries, 8,968,524 crime instances were retained, accounting for the loss of 482,503 instances, which corresponds to approximately $\sim$5.11\% of the total dataset. Following this, the crime instances were sorted chronologically and histogrammed. This curated form of the dataset was used as both training and test data for all neural network models.

Regarding the external factor information, which serves as an intermediate input for certain models, the data was segregated based on date and crime type before being histogrammed to produce a count of crime instances per crime type for each input coordinate given to the models. The intermediate inputs, which in other established spatiotemporal neural networks are interpreted as external factors, are not the canonical external factors used in GTWR. Instead of external independent variables, the intermediate inputs fed to some of the models in this section are the previous month's crime count of each crime type for the spatio-temporal coordinate given at the starting input layer. For example, the model layers labelled \textit{``EXTERNAL FACTORS (t)''} and \textit{``EXTERNAL FACTORS (t-1)''} refer to the
crime count of each crime type one and two months ago, respectively, for the input region in question. Due to difficulties in obtaining suitable quality data we adopted this scheme (as the raw data contained the crime type breakdowns at suitable spatial and temporal resolutions) to produce a proof-of-concept argument for the effective capabilities of the models. We justify this by noting that the relationship between past crime type counts and current total crime count is expected to be nonlinear yet harbour some extent of autocorrelation and partial autocorrelation. Hence, the network is tasked with extracting the usable remaining relationships from past crime type counts to predict the current total crime count. This mimicks the notion of using real-world external factor information which is expected to have an imperfect relevance to the target output, while still maintaining a strong relationship with the target output.

Additionally, the longitude and latitude coordinate reference system (EPSG:4326), representing a latitude/longitude coordinate system centered on the Earth's mass, was reformatted into the British National Grid coordinate reference system (EPSG:27700). This transformation involved using the transverse Mercator map projection algorithm to convert spatial information from longitude and latitude to meters. To mitigate the potential introduction of large weight values arising from extensive input values, spatial information was further scaled from metres to kilometres by dividing it by 1000.

The spatial resolution of the pre-processed data is defined by a grid size of 36 by 28 cells. This resolution was determined by selecting an initial grid size of $N$ by $N$, followed by calculating the minimum distance along both the vertical and horizontal spatial axes. To achieve approximately equal minimum distances on both axes, a scaling factor was computed. Subsequently, an integer ratio pair was chosen to best approximate the scaling factor. The selected pair of ratios was then scaled up to achieve an appropriate size. Further adjustments were made to the number of vertical and horizontal bin sizes through iterative testing until the minimum distances along both axes were closely matched. In the case of the London crime dataset, each cell corresponds to a square area of approximately 1.56 km $\times$ 1.56 km.

\subsection{Detroit Crime Data Pre-processing}

The Detroit crime dataset utilised in this later subsections of research was collected from the archives of https://data.detroitmi.gov/datasets/rms-crime-incidents. The dataset covers a duration of 6 years, spanning from January 2017 to December 2022, and offers a daily temporal resolution, providing daily records of reported crime incidents. The spatial representation of the dataset adopts a longitude and latitude format, capturing spatial coordinates corresponding to street-level locations of reported crimes.

It should be noted that the Detroit crime dataset did not display missing entries or mislabeled regions, thus requiring no removal of entries. The data set comprises a total of 486,203 crime instances. These instances were arranged chronologically and histogrammed to form the training and test data sets for all the neural network models used in this study.

External factor information, which serves as input for specific models, was generated by segregating the data based on date and crime type before undergoing histogramming.

Similar to the pre-processing approach used for the London crime dataset, the longitude and latitude coordinate system (EPSG:4326) was transformed into meters using the world geodetic system (WGS 84) (EPSG:32617). Finally, the spatial information was further scaled from metres to kilometres by dividing it by 1000,  following a similar rationale described in the previous subsection for London crime.

The spatial resolution of the pre-processed Detroit crime data is represented by a grid size of 25 by 31 cells. The determination of this grid size followed a similar approach to that used for London crime data. Special consideration was given to ensuring that the final geographic cell size closely resembled the cell size used for the London data. Consequently, each cell in the grid corresponds to a square area of approximately 1.47 km $\times$ 1.45 km.

\subsection{Further Pre-processing Considerations: Data Augmentation}

Enhancing the accuracy of an artificial neural network (ANN) often involves acquiring more data, as it is a straightforward and effective strategy. However, in cases where obtaining additional data is challenging or impractical, ANN practitioners, particularly in the domain of image processing, may resort to employing data augmentation. This technique aims to artificially expand the dataset by generating new instances through transformations or modifications of existing data samples. To achieve success with data augmentation, it is crucial to ensure that the essential underlying patterns necessary for mapping the input to the output are reasonably preserved during the augmentation process. 

In our study, we investigated whether crime occurrences in London exhibit an isotropic spatial relationship. Mathematically, we sought to determine whether crime can be effectively modeled as a function of the distance between cells. To explore this concept, we observed that all cell-to-cell distances can be conserved by employing transformations derived from the $D_4$ dihedral symmetry group. We note that distances are conserved and a new map is produced under the following transformations:

\begin{itemize}
    \item 3 rotations: Rot(90), Rot(180), Rot(270)
    \item 4 reflections: Vertical, Horizontal, Diagonal, Off-diagonal
\end{itemize}

By incorporating these transformations, including the original data, we can potentially augment our dataset to eight times its original size, assuming that London crime adheres to an isotropic spatial relationship. The existence of eight symmetries can be demonstrated by assigning labels to the four corners of a map, e.g.:

\begin{figure}[H]
  \centering
  \includegraphics[width=0.35\linewidth]{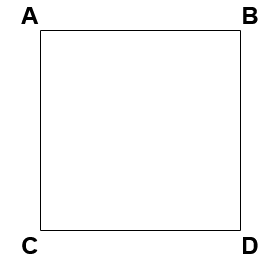}
\end{figure}

and through organising the labels into a set, we can confirm that cycling and reversing the labels results in distinct and adjacent sets.

\begin{align*}
    \overbrace{\{A,B,C,D\}}^{\textrm{Original set}} &\rightarrow \{D,A,B,C\} \rightarrow \{C,D,A,B\} \rightarrow \{B,C,D,A\} \\
    \overbrace{\{D,C,B,A\}}^{\textrm{Reversed set}} &\rightarrow \{A,D,C,B\} \rightarrow \{B,A,D,C\} \rightarrow \{C,B,A,D\}
\end{align*}

Consequently, there are eight symmetries that preserve distance on a map.

In practice we applied these transformations by combining the transformations of transposition and rotation, as follows:

\begin{itemize}
    \item Rot(90) = Rot(90)
    \item Rot(180) = Rot(180)
    \item Rot(270) = Rot(270)
    \item Vertical mirror = Transpose + Rot(90)
    \item Horizontal mirror = Transpose + Rot(270)
    \item Diagonal mirror = Transpose
    \item Off-diagonal mirror = Transpose + Rot(180)
\end{itemize}

\begin{figure}[H]
  \centering
  \includegraphics[width=0.5\linewidth]{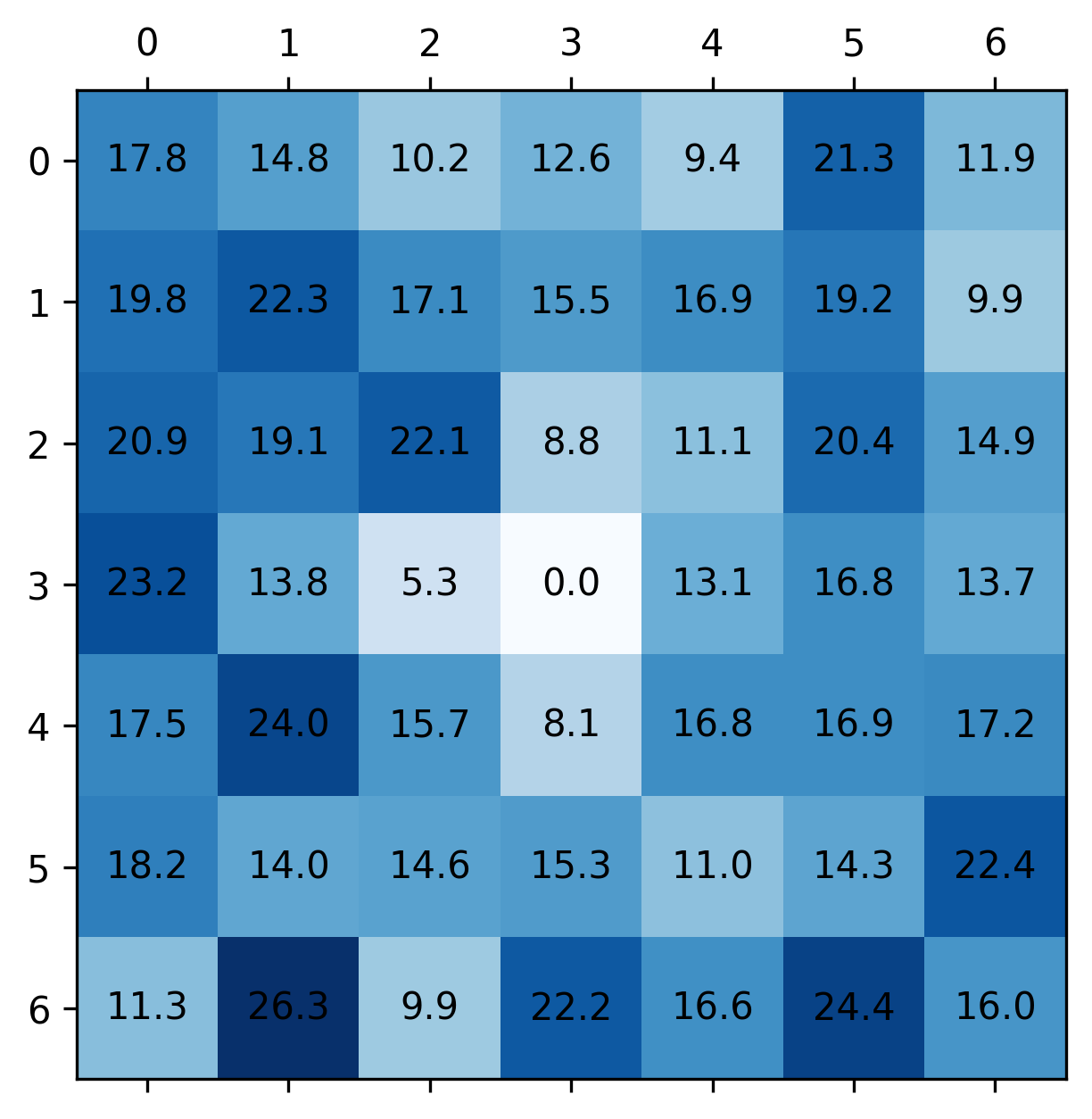}
  \caption{Aggregated Spatial Correlation Grid for London crime.}
  \label{isotropic_corr}
\end{figure}

To assess the viability of applying the aforementioned transformations and treating the new data on par with the original, we conducted a comprehensive analysis involving all potential cells. Specifically, we calculated the correlation between each cell and its neighboring cells within a 7x7 grid. In order to mitigate artifacts arising from overlapping grids, we randomly selected 25\% of these correlations and aggregated them (additively) to obtain a composite representation, as illustrated in \textcolor{blue}{\textbf{Figure \ref{isotropic_corr}}}.

In the context of isotropic treatment of crime occurrences, we would anticipate the presence of the symmetries mentioned earlier to be reflected in the correlation profile. However, upon inspection of the generated image, such symmetries are notably absent. Consequently, we can deduce that London crime does not exhibit an isotropic spatial relationship, and it is therefore inappropriate to apply data augmentation using the aforementioned transformations.

\subsection{Model Experiments}

\subsubsection{Training Parameters}

For all the models in this study, the test data comprised the final year of the dataset, while the remaining data was allocated for training. Both the training and test datasets were shuffled before they were passed to the network for training and validation.

Following the approach described in \cite{feng2021geographically}, certain parameters were fixed across all models to enable fair comparison:

\begin{itemize}
    \item The activation function for any neuron was set to ReLU($x$).
    \item The batch size was set to 10.
    \item The number of training epochs was set to 6.
    \item The optimisation algorithm used was adaptive moment estimation (ADAM) \citep{kingma2014adam}.
\end{itemize}

On the other hand, certain parameters were allowed to vary across the models, namely:

\begin{itemize}
    \item The number of hidden layers
    \item The number of neurons in a hidden layer
\end{itemize}

In reference to \textcolor{blue}{\textbf{Figures \ref{vanillaNNdiag}}} to \textcolor{blue}{\textbf{\ref{HDGTWNN_Lstdiag}}} of section 5.4,  these variable parameter choices influence the inner structure of the hidden layer blocks, which are denoted by the over-brace label \textit{``N Hidden Layers''} in the respective figures.

The search space for models without an intermediate input layer (Vanilla NN \& GWANN) was defined as follows:

\begin{itemize}
    \item The number of hidden layers $\in$ [1, 5] 
    \item The number of neurons in a hidden layer $\in$ [1, 15]
\end{itemize}

For models with at least one intermediate input layer, the search space was defined as:

\begin{itemize}
    \item The number of hidden layers $\in$ [1, 3] 
    \item The number of neurons in a hidden layer $\in$ [1, 15]
\end{itemize}

The range [1, 3] was chosen for models with at least one intermediate input layer to maintain a similar degree of non-linear complexity as the previous networks. Since models without an intermediate input layer could search up to 5 hidden layers, models with a single intermediate input layer were allowed to search up to 3 hidden layers, as they contain 2 hidden layer blocks, resulting in a total of 6 hidden layers. For history dependent models, even though they consist of 3 hidden layer blocks, the same search space was used. The rationale behind this decision is that the history dependent module serves as an extension to GTWNN, effectively duplicating the first portion of the network (excluding the first input layer). As both hidden layer blocks serve the purpose of producing influence factor values, it is justifiable to afford the history dependent module the same processing capabilities by keeping the search space consistent.

We opted for Bayesian optimisation as our chosen Neural Architecture Search (NAS) scheme for exploring the search space. The rationale behind this selection is elaborated upon in section 5.7.2.

\subsubsection{Bayesian Optimisation for Neural Architecture Search (NAS)}

The capability of how well a model can map an input to an output strongly depends on the architectural structure of the model. Large ANN models with many hidden layers and nodes (in each hidden layer) have the capacity to discern more complex patterns. While they have the capacity to discern more complex patterns, a larger network typically requires more data to become well trained. Comparatively, a small ANN model is expected to have a decreased capacity for detecting more sophisticated patterns. Given data is not in abundance, the practitioner must consider the trade-off between the balance of model complexity and simplicity. The fundamental notion concerning the appropriate scale of a neural model is examined and tackled within the domain of Neural Architecture Search (NAS). For the machine learning practitioner, there exist eight well-established strategies in the field of Neural Architecture Search (NAS) that warrant consideration. Among these eight, four are characterised by their relative simplicity in design, while the remaining four are distinguished by their more intricate nature. In this context, we present a concise overview of each strategy, with the purpose of elucidating the rationale behind the selection of Bayesian optimisation as the NAS approach for the current investigation.

\begin{table}[H]
\centering
\begin{tabular}{p{5.5cm}p{9cm}}
\multicolumn{2}{c}{\Large \textbf{Simple NAS Strategies}}  \vspace{2mm}                                                                                                                                                                \\ \hline
\\
\textbf{Grid search:} & Exhaustively explores all hyperparameter combinations, training each network separately.\vspace{3mm}                                                                                                       \\ 
\textbf{Random search:}         & Selects hyperparameter combinations randomly and trains the network. Repeats for $N$ iterations.\vspace{3mm}                                                                                     \\
\textbf{Hyperband:}             & Gradually reduces the number of configurations by halving them, distributing epochs randomly to the remaining configurations for each round.\vspace{3mm}  \\
\textbf{Bayesian Optimisation:} & Employs Gaussian process models to determine the next set of hyperparameters, maximising the potential improvement based on past experiments.                                                    
\end{tabular}
\caption{Brief description of the four common basic neural architecture search (NAS) strategies.}
\end{table}

\begin{table}[H]
\centering
\begin{tabular}{p{5.5cm}p{9cm}}

\multicolumn{2}{c}{\Large \textbf{Advanced NAS Strategies}}
\vspace{2mm} \\ \hline
\\
\textbf{NAS:} & utilises an RNN (Recurrent Neural Network) output to configure a sequence of tokens which make a cell/layer. These cells will either keep the dimensionality of the input or halve it. They are then stacked on top of one another to make a child network which is trained. The accuracy is fed into the RNN's reinforcement learning loss function and retrained.\vspace{3mm}                                                                                                        \\
\textbf{NAS without training:}         & Computes the correlation of the Jacobian which finds the tangent of the loss at each input datapoint for a given \textbf{untrained} configuration. Low correlation means that the majority of the networks components are doing an independent job. The best networks tend to have this property.\vspace{3mm}                                                                                     \\
\textbf{Differentiable learning:}             & Send input into every layer in the search space. Multiply each output by $\zeta_i \in [0,1]$ with $\sum_{i} \zeta_i = 1$ then add them together. Feed output to every search space layer again and multiply by variables in the same way. These variables are trained with the network and after you train the stacked network you pick the layers with the highest coefficients to construct the best network.\vspace{3mm}  \\
\textbf{Regularised evolution:} & Use a genetic algorithm to construct a population of networks by adding together $N$ randomly selected layers in your search space. Train. Out of those randomly pick a sample subset, pick a few of the best performing ones from the subset. Mutate them to make new networks. Repeat until convergence.                                                   
\end{tabular}
\caption{Brief description of the four common advanced neural architecture search (NAS) strategies.}
\end{table}
In practical applications, the advanced strategies of Neural Architecture Search (NAS), differentiable learning, and regularised evolution are commonly employed when the practitioner needs to explore diverse layer types or modules (e.g., ResNet and Inception modules). In our specific case, we focus on varying two scalable parameters within our search space: the number of hidden layers and the number of nodes independently within each hidden layer. We intentionally limit the scope of parameter variation to these aspects to ensure a fair comparison, as these are the only parameters investigated in the SOTA network (GTWNN) paper.

When these techniques are tasked with varying scalable parameters, they are required to incorporate all possible integer numbers of nodes (within a specified range) in the search space as independent layer types. This approach can lead to performance issues since these techniques are not inherently well-suited for handling such extensive configurations. Notably, NAS without training demonstrates a capacity to manage scalable parameters; however, despite its remarkable speed, it is primarily used to identify and eliminate degenerate configurations. Consequently, it does not consistently converge towards a specific optimal configuration due to the impracticality of reasonably training every possible network configuration, which incurs significant computational costs.

In the context of simple NAS strategies, Grid search emerges as the most computationally demanding method, often necessitating a carefully constrained search space to be practically applicable. Conversely, Random search does not face this limitation. However, its efficacy in exploring configurations does not improve over time, leading to a time-consuming and somewhat arbitrary process of generating reasonable configurations. Random search becomes a favorable choice when the search space incorporates categorical variables.

Hyperband theoretically offers the potential for improving network configuration as time progresses. Nevertheless, in practice, it also exhibits a degree of randomness, as a significant number of configurations are pruned before reaching a reasonable level of convergence. Consequently, promising network configurations might be prematurely eliminated.

Bayesian optimisation, on the other hand, samples the search space strategically using Gaussian likelihoods, targeting under-explored regions and assessing their worth for further exploration. This approach exhibits progressive improvement over time. Notably, Bayesian optimisation excels in handling scalable parameters, as it leverages an acquisition function to discern crucial regions in the search space for focused exploration.

However, it should be acknowledged that Bayesian optimisation encounters challenges when categorical parameters are involved, requiring a more exhaustive search due to the inability to assess their numerical similarity.

Given these considerations, Bayesian optimisation was selected as the preferred approach for this study, owing to its ability to adapt and improve over time while effectively handling scalable parameters.

For each Artificial Neural Network (ANN) model presented in this section, we conducted a search for 50 configurations utilising Bayesian optimisation after defining the respective search space. The rationale behind selecting this specific number of configurations is founded on the following argument:

If we were to randomly select a configuration, the probability of it belonging to the top 5\% of configurations is 5\%, leaving a 95\% chance of it not being among the top 5\%. Consequently, the likelihood of not encountering a network within the top 5\% of configurations after $n$ random selections can be expressed as $(0.95)^n$. Conversely, the probability of discovering a network within the top 5\% of configurations after $n$ random selections can be formulated as follows:

\begin{equation*}
1 - (0.95)^n
\end{equation*}

Solving the inequality:

\begin{equation}
1 - (0.95)^n > 0.95, \quad n \in \mathbb{Z}^+
\end{equation}

yields $n \geq 60$. Therefore, with $n=60$ random selections, there exists a probability above 95\% of finding a network within the top 5\% of configurations. Considering that Bayesian optimisation conducts its search strategically rather than randomly, the likelihood of successfully identifying a configuration in the top 5\% is expected to be superior to random chance. Consequently, in practice, many practitioners have observed that a more modest number of $n=50$ configurations suffices for Bayesian optimisation to perform effectively.

\subsection{London Crime Model Results}

In this section, we present the results for each model individually, following the order of their appearance in section 5.4. We will provide a separate discussion for the results of each model, analysing them independently. Subsequently, we will present the collective results, which consolidate the best model outcomes for comparison and evaluation of the models against each other.

\subsubsection{Vanilla NN Results for London Crime}

The following table displays the achieved evaluation metric scores, including MSE, MAPE, and R2, resulting from the implementation of the vanilla neural network (NN) model on the London crime dataset.

\begin{table}[H]
\doublespacing
\centering
\begin{tabular}{llll}
\hline
\multicolumn{4}{c}{\textbf{Vanilla NN results for London Crime}} \\ \hline
                  & \textbf{MSE}         & \textbf{MAPE}        & \textbf{R2}      \\
1 Hidden Layer    & 36187.652   & 4326726656  & 0.0997  \\
2 Hidden Layers   & 34943.305   & 3898339072  & 0.131   \\
3 Hidden Layers   & 35179.297   & 4338387968  & 0.125   \\
4 Hidden Layers   & 35747.781   & 4146821376  & 0.111   \\
5 Hidden Layers   & 37456.172   & 4952271872  & 0.068  
\end{tabular}
\caption{Vanilla NN results for London crime.}
\end{table}

It is worth noting that the vanilla neural network (NN) yields high Mean Squared Error (MSE) scores, significantly large Mean Absolute Percentage Error (MAPE) scores, and an R2 coefficient of determination close to zero. As the vanilla NN constitutes the most elementary model among those presented, lacking a customised structure for spatiotemporal problems, it is anticipated that all subsequent models should generally exhibit improved metric scores compared to the vanilla NN.

\subsubsection{GWANN Resutls for London crime}

The following table displays the achieved evaluation metric scores, including MSE, MAPE, and R2, resulting from the implementation of the GWANN model on the London crime dataset.

\begin{table}[H]
\doublespacing
\centering
\begin{tabular}{cccc}
\hline
\multicolumn{4}{c}{\textbf{GWANN Results for London Crime}}  \\ \hline
                & \textbf{MSE} & \textbf{MAPE} & \textbf{R2} \\
1 Hidden Layer  & 37,408.311   & 0.681         & -1.427      \\
2 Hidden Layers & 36,428.150   & 0.757         & -1.009      \\
3 Hidden Layers & 36,848.859   & 0.654         & -0.904      \\
4 Hidden Layers & 36,013.338   & 0.720         & -1.127      \\
5 Hidden Layers & 36,081.522   & 0.833         & -0.793     
\end{tabular}
\caption{GWANN results for London crime.}
\end{table}

We observe that the GWANN model exhibits only marginal enhancements in MSE when compared to the vanilla NN model, but it demonstrates a significant improvement in MAPE. However, the R2 coefficient remains notably poor. The enhanced MAPE implies improved predictions for spatiotemporal locations with very few to zero observations. The limited improvement and the expected advantage of GWANN indicate a potential deficiency in the spatial correlation of neighbouring cells within the dataset, either entirely or at the chosen spatial resolution for the cells.

\subsubsection{GTWNN Results for London Crime}

The following table displays the achieved evaluation metric scores, including MSE, MAPE, and R2, resulting from the implementation of the GTWNN model on the London crime dataset.

\begin{table}[H]
\doublespacing
\centering
\begin{tabular}{cccc}
\hline
\multicolumn{4}{c}{\textbf{GTWNN results for London Crime}}  \\ \hline
                & \textbf{MSE} & \textbf{MAPE} & \textbf{R2} \\
1 Hidden Layer  & 1306.569     & 364388704     & 0.858       \\
2 Hidden Layers & 1061.1869    & 176984000     & -2.085      \\
3 Hidden Layers & 1494.322     & 353040384     & 0.784      
\end{tabular}
\caption{GTWNN results for London crime.}
\end{table}

Among the evaluation scores, we observe a significant improvement in MSE and R2. However, it is essential to note that obtaining a good R2 score seems highly dependent on the specific architecture chosen. Therefore, when employing this model, cautious selection of the NAS strategy is advised. The noteworthy R2 scores achieved through Bayesian optimisation, with 3 and 1 hidden layer(s) in each hidden block, indicate that GTWNN effectively captures major trend patterns observed in London crime. Nonetheless, the remarkably high MAPE score implies that the model encounters difficulties in predicting future observations that have zero values.

\subsubsection{GTWNN\_Ls Results for London Crime}

The following table displays the achieved evaluation metric scores, including MSE, MAPE, and R2, resulting from the implementation of the GTWNN\_Ls model on the London crime dataset.

\begin{table}[H]
\doublespacing
\centering
\begin{tabular}{cccc}
\hline
\multicolumn{4}{c}{\textbf{GTWNN\_Ls Results for London Crime}} \\ \hline
                      & \textbf{MSE}      & \textbf{MAPE}      & \textbf{R2}      \\
1 Hidden Layer        & 27,623.619        & 0.225              & 0.075            \\
2 Hidden Layers       & 26,656.049        & 0.265              & -0.164           \\
3 Hidden Layers       & 26,424.881        & 0.229              & -0.055          
\end{tabular}
\caption{GTWNN\_Ls Results for London crime.}
\end{table}

For the GTWNN\_Ls model, we observe a reduction in MSE compared to GWANN, indicating that the incorporation of the intermediate layer linking the neural network's architecture to the theory of GTWR imparts enhanced predictive capabilities to the overall model. However, despite this improvement, the obtained MSE scores remain notably inferior to those observed in GTWNN. This suggests that neighboring cells at the current resolution exhibit low spatial correlation. On the other hand, the MAPE score for GTWNN\_Ls is the lowest among the individual model results presented thus far, indicating that the model can effectively handle predictions for future instances with zero and low crime counts. Nonetheless, the R2 score is nearly zero, implying that the enforcement of spatial continuity in the estimation of the $\beta$ functions hinders the network's ability to discern the general trend in crime. This further supports the notion of low spatial correlation among neighboring cells in the dataset at the given cell size.

\subsubsection{GTWNN\_Lst Results for London Crime}

The following table displays the achieved evaluation metric scores, including MSE, MAPE, and R2, resulting from the implementation of the GTWNN\_Lst model on the London crime dataset.

\begin{table}[H]
\doublespacing
\centering
\begin{tabular}{cccc}
\hline
\multicolumn{4}{c}{\textbf{GTWNN\_Lst Results for London Crime}} \\ \hline
                                & \textbf{MSE}                 & \textbf{MAPE}                 & \textbf{R2}                \\
1 Hidden Layer                  & 21,503.202                   & 0.424                         & 0.281                      \\
2 Hidden Layers                 & 22,409.152                   & 0.363                         & 0.312                      \\
3 Hidden Layers                 & 21,863.171                   & 0.418                         & 0.211                     
\end{tabular}
\caption{GTWNN\_Lst results for London crime.}
\end{table}

In this context, it is noteworthy that extending the output layer both spatially and temporally, as opposed to solely spatially (as in the case of GTWNN\_Ls), leads to improved MSE and R2 scores. This suggests that there exists sufficient temporal correlation among neighboring cells in the dataset, making this strategy advantageous. However, it should be mentioned that the MAPE has increased, indicating a reduced ability to handle predictions for future instances with low or zero crime occurrences. Nevertheless, the increase in MAPE is not significantly concerning, given the observation that the R2 has increased to a meaningful level. This implies that the model has captured some of the general trends in London crime, and the inclusion of temporal expansion in the output layer has contributed to this improvement.

\subsubsection{HDGTWNN Results for London crime}

The following table displays the achieved evaluation metric scores, including MSE, MAPE, and R2, resulting from the implementation of the HDGTWNN model on the London crime dataset.

\begin{table}[H]
\doublespacing
\centering
\begin{tabular}{cccc}
\hline
\multicolumn{4}{c}{\textbf{HDGTWNN results for London Crime}} \\ \hline
                 & \textbf{MSE} & \textbf{MAPE} & \textbf{R2} \\
1 Hidden Layer   & 1248.072     & 5512214.5     & 0.932       \\
2 Hidden Layers  & 1140.089     & 1587972.5     & 0.947       \\
3 Hidden Layers  & 1100.388     & 1581844.625   & 0.940      
\end{tabular}
\caption{HDGTWNN results for London crime.}
\end{table}

When comparing HDGTWNN to GTWNN, it is evident that the MSE shows minimal variation. However, notable improvements are observed in the MAPE and a consistent enhancement in the R2 value. HDGTWNN achieves the highest R2 score among all the individual model results presented thus far. These improvements suggest that the incorporation of the history dependent module leads to better predictions for lower crime instances and enhances the model's capability to capture the general trends in London crime.

\subsubsection{HDGTWNN\_Ls Results for London crime}

The following table displays the achieved evaluation metric scores, including MSE, MAPE, and R2, resulting from the implementation of the HDGTWNN\_Ls model on the London crime dataset.

\begin{table}[H]
\doublespacing
\centering
\begin{tabular}{cccc}
\hline
\multicolumn{4}{c}{\textbf{HDGTWNN\_Ls Results for London Crime}} \\ \hline
                               & \textbf{MSE}                & \textbf{MAPE}               & \textbf{R2}               \\
1 Hidden Layer                 & 25,814.180                  & 0.210                       & -0.0277                   \\
2 Hidden Layers                & 23,061.654                  & 0.184                       & -0.0374                   \\
3 Hidden Layers                & 25,789.289                  & 0.245                       & -0.0286                  
\end{tabular}
\caption{HDGTWNN\_Ls results for London crime.}
\end{table}

We begin by observing that HDGTWNN\_Ls exhibits the lowest MAPE among the models presented thus far. In comparison to GTWNN\_Ls, both the MSE and MAPE have decreased, suggesting that the inclusion of the history dependent module contributes to a general enhancement in the model's predictive capability. However, the R2 value remains approximately at zero correlation, indicating that while there is an improvement in the predictive ability of the network, it has not captured the general crime trends. This finding suggests a high degree of temporal correlation and low spatial correlation between neighboring cells in the dataset.

\subsubsection{HDGTWNN\_Lst Results for London crime}

The following table displays the achieved evaluation metric scores, including MSE, MAPE, and R2, resulting from the implementation of the HDGTWNN\_Lst model on the London crime dataset.

\begin{table}[H]
\doublespacing
\centering
\begin{tabular}{cccc}
\hline
\multicolumn{4}{c}{\textbf{HDGTWNN\_Lst Results for London Crime}} \\ \hline
                                & \textbf{MSE}                 & \textbf{MAPE}                 & \textbf{R2}                \\
1 Hidden Layer                  & 25,918.644                   & 0.218                         & 0.0566                     \\
2 Hidden Layers                 & 25,367.511                   & 0.287                         & -0.0757                    \\
3 Hidden Layers                 & 23,623.176                   & 0.255                         & 0.0324                    
\end{tabular}
\caption{HDGTWNN\_Lst results for London crime.}
\end{table}

The outcomes obtained for HDGTWNN\_Lst display marginal differences when compared to HDGTWNN\_Ls. This observation suggests that the history dependent module can effectively leverage the temporal correlation among neighboring cells in the dataset, rendering the inclusion of temporal expansion in the output layer redundant. Consequently, when faced with the option of either expanding the output layer temporally or incorporating a history dependent module, the latter proves to be the more advantageous choice.

\subsubsection{Aggregated Best Model Results for London Crime}

Here, we present the aggregated results for all the ANN models applied to the London crime dataset. Each entry in the table represents the lowest MSE score obtained from our NAS strategy, accompanied by their corresponding MAPE and R2 scores. The models are arranged in descending order based on their MSE values.

\begin{table}[H]
\doublespacing
\centering
\begin{tabular}{cccc}
\hline
\multicolumn{4}{c}{\textbf{Aggregated Best Model Results for London Crime}} \\ \hline
\textbf{}      & \textbf{MSE}   & \textbf{MAPE}   & \textbf{R2}  \\
GWANN          & 36,013.338     & 0.720           & -1.127       \\
Vanilla NN     & 34943.305      & 3898339072      & 0.131        \\
GTWNN\_Ls      & 26,424.881     & 0.229           & -0.055       \\
HDGTWNN\_Lst   & 23,623.176     & 0.255           & 0.0324       \\
HDGTWNN\_Ls    & 23,061.654     & 0.184           & -0.0374      \\
GTWNN\_Lst     & 21,503.202     & 0.424           & 0.281        \\
HDGTWNN        & 1100.388       & 1581844.625     & 0.940        \\
GTWNN          & 1061.1869      & 176984000       & -2.085      
\end{tabular}
\caption{Aggregated best model results for London crime arranged in order of decreasing MSE scores.}
\end{table}

GTWNN achieved the best MSE score, with HDGTWNN closely following suit. HDGTWNN\_Ls demonstrated the lowest MAPE among all models. HDGTWNN obtained the best R2 score. These findings indicate that incorporating the history dependent module does not compromise the benefits of the models lacking such a module. Overall, the results suggest that the history dependent module consistently contributes advantageous enhancements to the predictive capabilities of the network.

Conversely, GWANN exhibits the worst MSE scores, closely followed by vanilla NN. These results imply that the inclusion of a spatially expanded output layer diminishes predictive power, indicating a low spatial correlation between neighboring cells in this dataset. Generally, models attempting to leverage the spatial correlation of neighboring cells to enhance predictive capabilities have performed less effectively for the London crime dataset.

To some readers, some of the metric values presented in \textcolor{blue}{\textbf{Table 18}} seem concerningly large (MAPE) or difficult to interpret (negative R2 values). While some suggested logic has been presented in earlier subsections to explain this we attempt to present a clearer picture for these metric scores in the following subsection by visually analysing the outputs of GTWNN on the London crime dataset as it contains a large MAPE and negative R2 score.

\subsubsection{Visual Analysis of GTWNN Output for London Crime Data}

A negative R2 score refers to a model which underperforms against using the time averaged mean as the model for all predictions. When concerning the results in \textcolor{blue}{\textbf{Table 18}} we must firstly note that these scores are derived from the model operating on unseen test set data. So, a negative R2 score refers to a model underperforming against the time-averaged test set mean. When visually comparing the actual average crime distribution of the test set against the average crime distribution for the test set predicted by the GTWNN model we produce:

\begin{figure}[H]
\makebox[\linewidth]{
\begin{subfigure}{0.45\textwidth}
  \centering
  \includegraphics[width=1.1\linewidth]{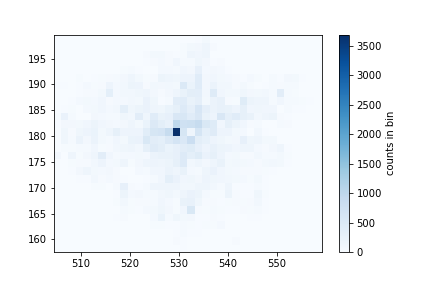} 
  \caption{The time-averaged crime distribution for the test set.}
  \label{fig5:1}
\end{subfigure}\hfil 
\begin{subfigure}{0.45\textwidth}
  \centering
  \includegraphics[width=1.1\linewidth]{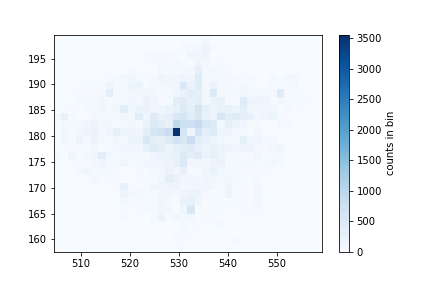}
  \caption{The model predicted time-averaged crime distribution for the test set.}
  \label{fig5:2}
\end{subfigure} 
}
\caption{Visual comparison of the actual and predicted time-averaged crime distribution for the test set.}
\label{avg1}
\end{figure}

Upon initial inspection, the model suggested time-averaged crime distribution appears quite similar to the actual time-averaged crime distribution. By plotting the difference between the actual and predicted time-averaged crime distributions we can determine if and where crime count is generally over and under estimated by the model:

\begin{figure}[H]
  \centering
  \includegraphics[width=0.5\linewidth]{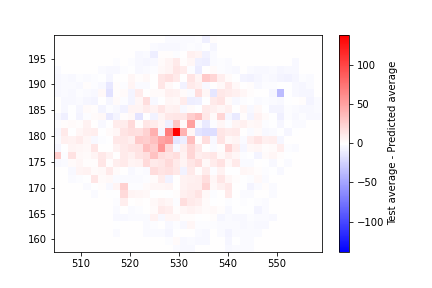}
  \caption{Difference between the actual and model predicted time-averaged crime distributions for the test set.}
  \label{diff1}
\end{figure}

A red grid cell in \textcolor{blue}{\textbf{Figure \ref{diff1}}} shows where the model is generally underestimating. Conversely, a blue cell shows where the model is generally overestimating. Some of these cells where the model overestimates are very close to zero crimes in the actual average. In total, of the 8364 possible inputs representing the test set, 386 were predicted to have crime count greater than 1 where the true crime count was zero, leading to very large MAPE values. 

To understand the negative R2 score we can compare the time-averaged crime distribution of the training set against the time-average crime distribution of the model's output when the training set is provided as the input:

\begin{figure}[H]
\makebox[\linewidth]{
\begin{subfigure}{0.45\textwidth}
  \centering
  \includegraphics[width=1.1\linewidth]{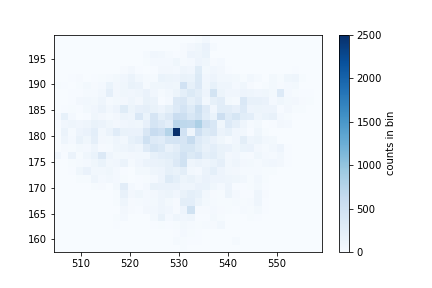} 
  \caption{The time-averaged crime distribution for the training set.}
  \label{fig5:3}
\end{subfigure}\hfil 
\begin{subfigure}{0.45\textwidth}
  \centering
  \includegraphics[width=1.1\linewidth]{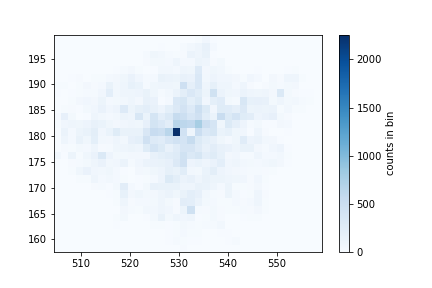}
  \caption{The model predicted time-averaged crime distribution for the training set.}
  \label{fig5:4}
\end{subfigure} 
}
\caption{Visual comparison of the actual and predicted time-averaged crime distribution for the training set.}
\label{fig5-vis}
\end{figure}

Initial inspections again suggest that the true observed and model's prediction of the time-averaged crime distribution for the training set are similar. However, differing from \textcolor{blue}{\textbf{Figure \ref{avg1}}}, the average number of crimes per grid cell is scaled down. This is concurrent with lower crime rates in the past. Taking the difference between the actual and predicted time-averaged crime distribution for the training set produces:

\begin{figure}[H]
  \centering
  \includegraphics[width=0.5\linewidth]{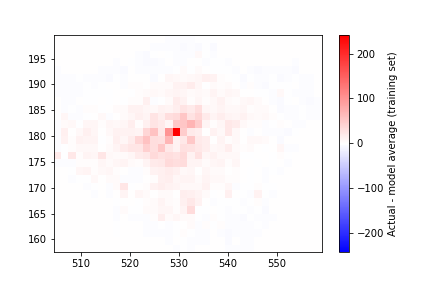}
  \caption{Difference between the actual and model predicted time-averaged crime distributions for the training set.}
  \label{diff2}
\end{figure}

Where we can see that the model prediction for the time-averaged training set crime distribution generally underestimates the crime count when the training set is provided as input. We hypothesise that the model relies too strongly on statistical properties present in the training set leading to over-estimation when the model is applied to unseen test data. As we have seen, the model is capable of producing a reasonably well-scaled distribution concurrent with the increased crime rate evolving with time. However, if we re-scale the time-averaged crime distribution for the training set such that the maximum crime count in a cell equates (effectively fitting the tail-end of the distribution), we can produce the difference between the two distributions:

\begin{figure}[H]
  \centering
  \includegraphics[width=0.5\linewidth]{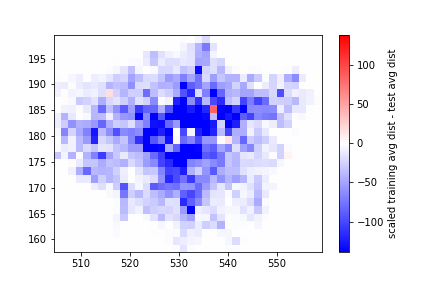}
  \caption{Difference between the (re-scaled) training and test time-averaged crime distributions.}
  \label{diff3}
\end{figure}

When using a re-scaled time-averaged crime distribution from the training set, it is clear (from \textcolor{blue}{\textbf{Figure \ref{diff3}}}) that a model relying too heavily on adopting statistical properties from the training set will generally lead to overestimations for the majority of spatial co-ordinates when the test set is provided as input. As such, it is likely that the model produces the overestimates seen in \textcolor{blue}{\textbf{Figure \ref{diff1}}} due to an over-reliance of learned statistical properties in the training data which do not generalize to the test data.

\subsection{Detroit Crime Dataset vs London Crime Dataset}

One large discrepancy between the London and Detroit crime datasets is that they are of different temporal resolutions. 

\begin{equation}
     x_t \approx x_{t-1} + \Delta t \frac{d(x_{t-1})}{dt}
     \label{starteq}
\end{equation}
where $x_t$ represents the value of $x$ at time $t$, $\Delta t$ represents the smallest time interval, and $\frac{d(x_{t-1})}{dt}$ is a notational shorthand form of:
\begin{equation*}
    \frac{dx(t)}{dt}\bigg|_{t=t-1}
\end{equation*}
This approximation becomes more accurate the smaller $\Delta t$ is. When $\Delta t$ is infinitesimally small (\ref{starteq}) becomes an equality:

\begin{equation}
     x_t = x_{t-1} + \lim_{\Delta t \rightarrow 0} \left[\Delta t \frac{d(x_{t-1})}{dt}\right]
     \label{equality}
\end{equation}

Using $\lim_{\Delta t \rightarrow 0} (\Delta t) = dt$ we can more concisely rewrite (\ref{equality}) as:

\begin{equation}
     x_t = x_{t-1} +  dt \frac{d(x_{t-1})}{dt}
\end{equation}

note that:
\begin{align*}
    \frac{d(x_{t-1})}{dt} &= \frac{d}{dt}\left(x_{t-2} + \Delta t \frac{d(x_{t-2})}{dt}\right) \\
    &= \frac{d(x_{t-2})}{dt} + \Delta t \frac{d^2(x_{t-2})}{dt^2} \\
    &= \frac{d(x_{t-3})}{dt} + \Delta t \frac{d^2(x_{t-3})}{dt^2} + \Delta t \frac{d^2(x_{t-2})}{dt^2}
\end{align*}
and in general:
\begin{equation}
    x_t \approx x_{t-1} + \sum_{i=1}^{t-1} \Delta t^2 \frac{d^2 x_i}{dt^2} + \Delta t^2 \frac{d x_1}{dt}
\end{equation}
or equivalently:
\begin{equation}
    x_{t+n} \approx x_{t+n-1} + \sum_{i=t}^{t+n-1} \Delta t^2 \frac{d^2 x_i}{dt^2} + \Delta t^2 \frac{d x_t}{dt}
\end{equation}
When comparing the London and Detriot dataset, the London dataset has a maximum temporal resolution of 1 month $\approx$ 30 days, and the Detriot dataset has a maximum temporal resolution of 1 day. Suppose we try to predict the crime count for next month, using a daily dataset, the second term is approximately:
\begin{equation}
   \sum_{i=t}^{t+30} \frac{d^2 x_i}{dt^2}
   \label{dayderiv}
\end{equation}
However, for a monthly dataset this is just:
\begin{equation}
    \frac{d^2 x_t}{dt^2}
    \label{monderiv}
\end{equation}
where $x_t$ represents the number of crimes
but, in the context of a daily dataset $\Delta t \approx 30$ and (\ref{monderiv}) can be re-represented as:
\begin{equation}
    (30)^2 \frac{d^2 x_t}{dt^2}
    \label{finderiv}
\end{equation}
equating (\ref{dayderiv}) and (\ref{finderiv}) we find that the following must hold true:
\begin{equation}
    \left\langle \frac{d^2 x_i}{dt^2} \right\rangle = (30)\frac{d^2 x_t}{dt^2}, \quad \forall i
\end{equation}
to produce a reasonable result, which is generally not the case. Hence when using a monthly dataset, there is information loss regarding the evolution of the system on smaller time-frames. Mathematically, a future prediction using a monthly dataset effectively applies the same daily evolution $\Delta t = 30$ times. Additionally, for historic data to be considered useful for time-series forecasting in any capacity it is imperative that future observations must at least depend on the current time point. Furthermore, the application of historic data to reasonably aid in the prediction of future observations can only occur if, and only if, future observations depend on historic observations to some accepted threshold of significance, i.e. that:
\begin{equation}
    p(x_{t+2},x_{t+1},x_{t}) = p(x_{t+2}|x_{t+1},x_t) p(x_{t+1}|x_t) p(x_t)
    \label{conditional}
\end{equation}
when future observations are not dependent on past observations (\ref{conditional}) regresses to:
\begin{equation}
    p(x_{t+2},x_{t+1},x_{t}) = p(x_{t+2}) p(x_{t+1}) p(x_t)
    \label{unconditional}
\end{equation}
where previous observations have no influence on future observations.

This implication implies that the historic data employed must as least harbour a certain acceptable level of (auto)correlation. The autocorrelation function, a statistical tool which providing the degree of correlation between observations separated by some time interval $\{ t-k,t \}$ is defined by:
\begin{equation}
    \rho(k) = \frac{\frac{1}{n-k}\sum_{t=k+1}^{N}(y_t - \bar{y})(y_{t-k} - \bar{y)}}{\sqrt{\frac{1}{n} \sum_{t=1}^{n} (y_t - \bar{y})^2} \sqrt{\frac{1}{n-k} \sum_{t=k+1}^{n} (y_{t-k} - \bar{y})^2}}
\end{equation}
Where $n$ is the sample size, and $k$ is the lag time. For the aforementioned models, the autocorrelation is statistically significant if it falls beyond the confidence bands defined by:
\begin{equation}
\label{confband}
    \pm z_{1-\alpha/2} \sqrt{\frac{1}{n}\left( 1 + 2\sum_{i=1}^{k} \rho(i)^2 \right)}
\end{equation}
where $z$ is the quantile function for the normal distribution, and $\alpha$ is the significance level. We fistly calculate the autocorrelation for both datasets, Detroit (daily granularity) and London (monthly granularity).
\begin{figure}[H]
\makebox[\linewidth]{
\captionsetup[subfigure]{oneside,margin={0.5cm,0cm}}
\begin{subfigure}{0.5\linewidth}
  \includegraphics[width=1.0\linewidth]{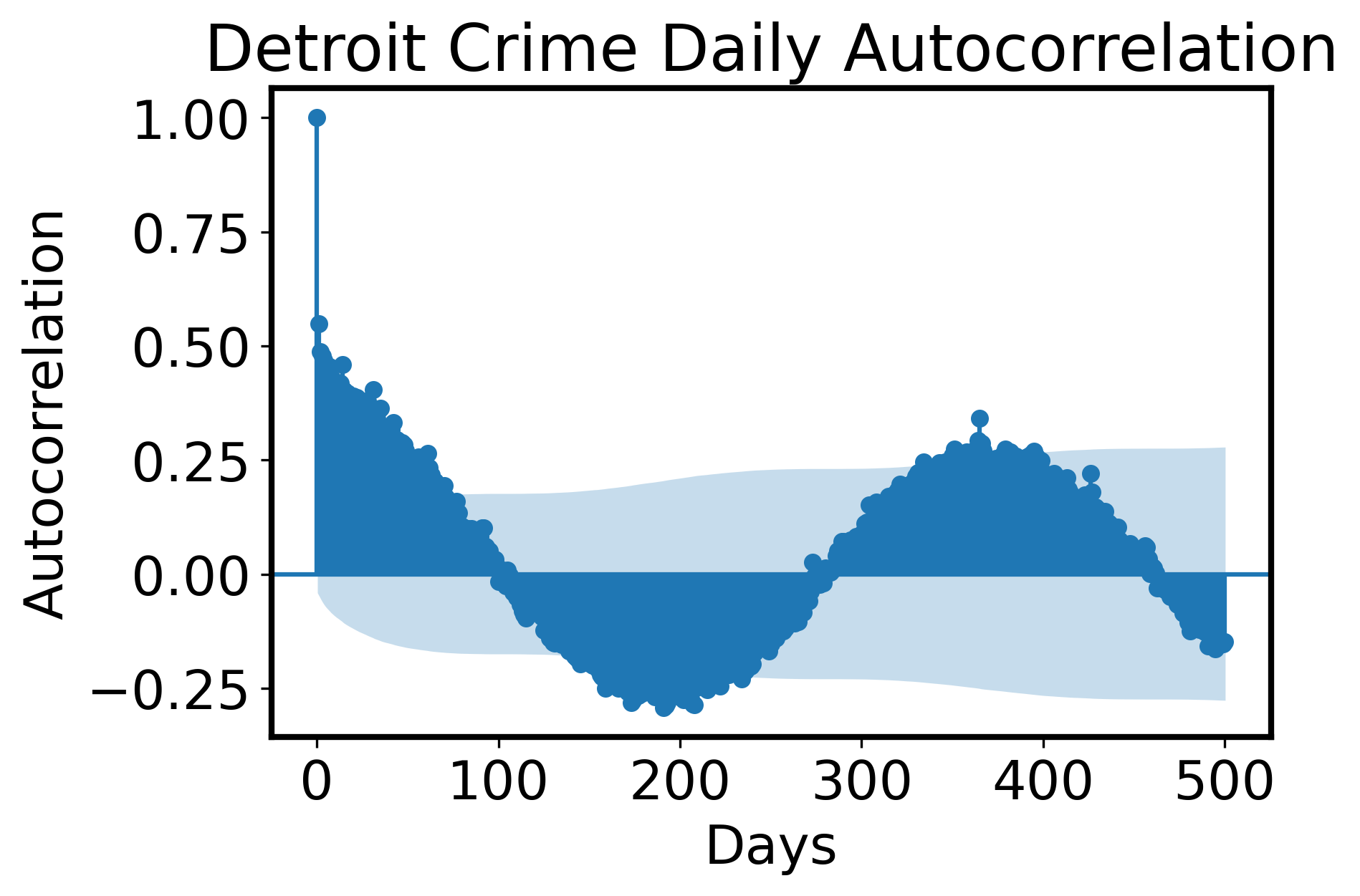} 
  \caption{Detroit daily auto-correlation over a 500 \\ day range.}
  \label{fig5:5}
\end{subfigure}\hfil 
\captionsetup[subfigure]{oneside,margin={0.5cm,0cm}}
\begin{subfigure}{0.5\linewidth}
  \includegraphics[width=1.0\linewidth]{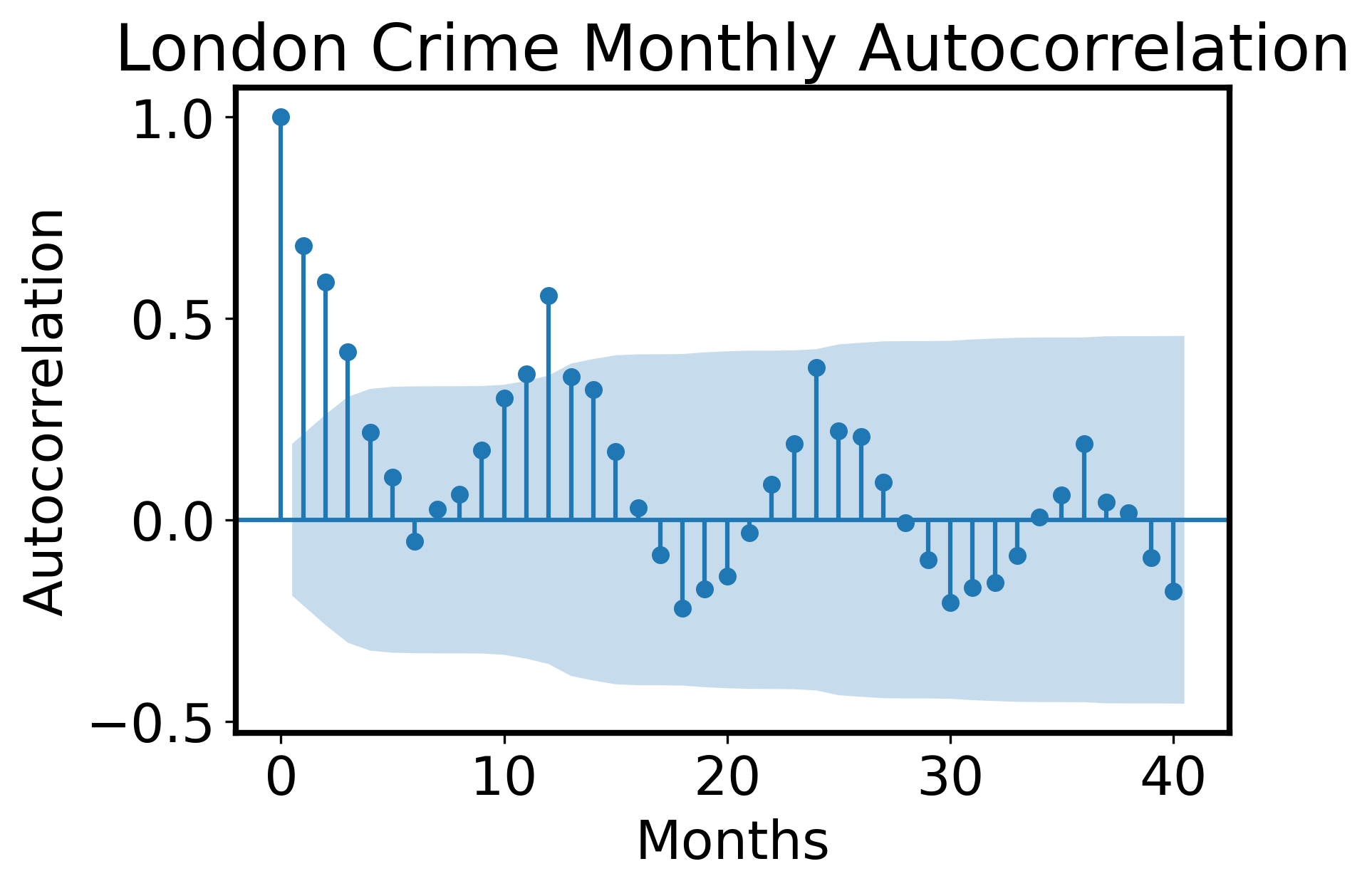}
  \caption{London monthly auto-correlation over a 40 \\ month range.}
  \label{fig5:6}
\end{subfigure} 
}
\caption{Autocorrelation for crime in London [monthly, \textbf{\textcolor{blue}{(a)}}] and Detroit [daily, \textbf{\textcolor{blue}{(b)}}] with shaded blue volume representing a 95\% statistical significance confidence interval.}
\label{acf}
\end{figure}
In both cases, the yearly periodicity is apparent, and the first 3 months ($\approx$90 days) both appear statistically significant. For the datasets to be applicable to the models we expect that the previous observation is statistically significant, and also the secondmost previous observation for history dependent models. While it would appear that both datasets seem applicable to the history dependent models, we have calculated both the direct and indirect contribution of the secondmost previous observation. 
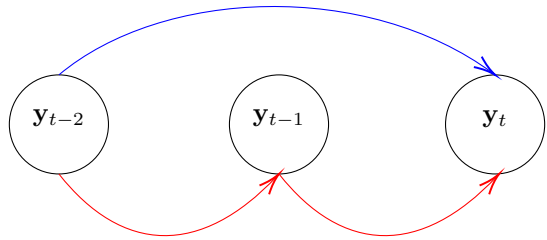
\begin{figure}[H]
\begin{center}
\begin{tikzpicture}[x=0.75pt,y=0.75pt,yscale=-1,xscale=1]

\draw   (177,208) .. controls (177,194.19) and (188.19,183) .. (202,183) .. controls (215.81,183) and (227,194.19) .. (227,208) .. controls (227,221.81) and (215.81,233) .. (202,233) .. controls (188.19,233) and (177,221.81) .. (177,208) -- cycle ;
\draw   (288,208) .. controls (288,194.19) and (299.19,183) .. (313,183) .. controls (326.81,183) and (338,194.19) .. (338,208) .. controls (338,221.81) and (326.81,233) .. (313,233) .. controls (299.19,233) and (288,221.81) .. (288,208) -- cycle ;
\draw   (397,208) .. controls (397,194.19) and (408.19,183) .. (422,183) .. controls (435.81,183) and (447,194.19) .. (447,208) .. controls (447,221.81) and (435.81,233) .. (422,233) .. controls (408.19,233) and (397,221.81) .. (397,208) -- cycle ;
\draw [color={rgb, 255:red, 255; green, 0; blue, 0 }  ,draw opacity=1 ]   (202,233) .. controls (233.98,273.59) and (273.5,274.98) .. (311.84,234.25) ;
\draw [shift={(313,233)}, rotate = 492.66] [color={rgb, 255:red, 255; green, 0; blue, 0 }  ,draw opacity=1 ][line width=0.75]    (10.93,-3.29) .. controls (6.95,-1.4) and (3.31,-0.3) .. (0,0) .. controls (3.31,0.3) and (6.95,1.4) .. (10.93,3.29)   ;

\draw [color={rgb, 255:red, 255; green, 0; blue, 0 }  ,draw opacity=1 ]   (313,233) .. controls (344.98,273.59) and (384.5,274.98) .. (422.84,234.25) ;
\draw [shift={(424,233)}, rotate = 492.66] [color={rgb, 255:red, 255; green, 0; blue, 0 }  ,draw opacity=1 ][line width=0.75]    (10.93,-3.29) .. controls (6.95,-1.4) and (3.31,-0.3) .. (0,0) .. controls (3.31,0.3) and (6.95,1.4) .. (10.93,3.29)   ;

\draw [color={rgb, 255:red, 3; green, 0; blue, 255 }  ,draw opacity=1 ]   (202,183) .. controls (254.04,139.22) and (357.26,135.04) .. (421.04,182.28) ;
\draw [shift={(422,183)}, rotate = 217] [color={rgb, 255:red, 3; green, 0; blue, 255 }  ,draw opacity=1 ][line width=0.75]    (10.93,-3.29) .. controls (6.95,-1.4) and (3.31,-0.3) .. (0,0) .. controls (3.31,0.3) and (6.95,1.4) .. (10.93,3.29)   ;

\draw (202,205) node    {$\mathbf{y}_{t-2}$};
\draw (313,205) node    {$\mathbf{y}_{t-1}$};
\draw (422,205) node    {$\mathbf{y}_{t}$};

\end{tikzpicture}
\caption{A direct correlation between $y_{t-2}$ and $y_t$ (\textbf{\textcolor{blue}{blue}}), and an indirect correlation between $y_{t-2}$ and $y_t$ through $y_{t-1}$. (\textbf{\textcolor{red}{red}})} \label{pacf}
\end{center}
\end{figure}
Since we are not concerned with the indirect correlation between $y_{t-2}$ and $y_t$ (red arrows in \textcolor{blue}{\textbf{Figure \ref{pacf}}}) we can calculate the partial autocorrelation function (PACF) to isolate the direct correlation of $y_{t-2}$ to $y_t$. This is calculated by:
\begin{equation}
    \phi(k,k) = \frac{\rho(k) - \sum_{t=1}^{k-1} \phi(k-1,t) \rho(k-t)}{1 - \sum_{t=1}^{k-1} \phi(k-1,t) \rho(k-t)}, \quad \forall k > 1
\end{equation}
where $\phi(k,k)$ is the partial autocorrelation for lag time $k$ with $\phi(1,1) = \rho(1)$. The PACF was applied to both datasets:
\begin{figure}[H]
  \captionsetup[subfigure]{singlelinecheck=false}
\makebox[\linewidth]{
\captionsetup[subfigure]{oneside,margin={0.5cm,0cm}}
\begin{subfigure}{0.5\linewidth}
  \includegraphics[width=1.0\linewidth]{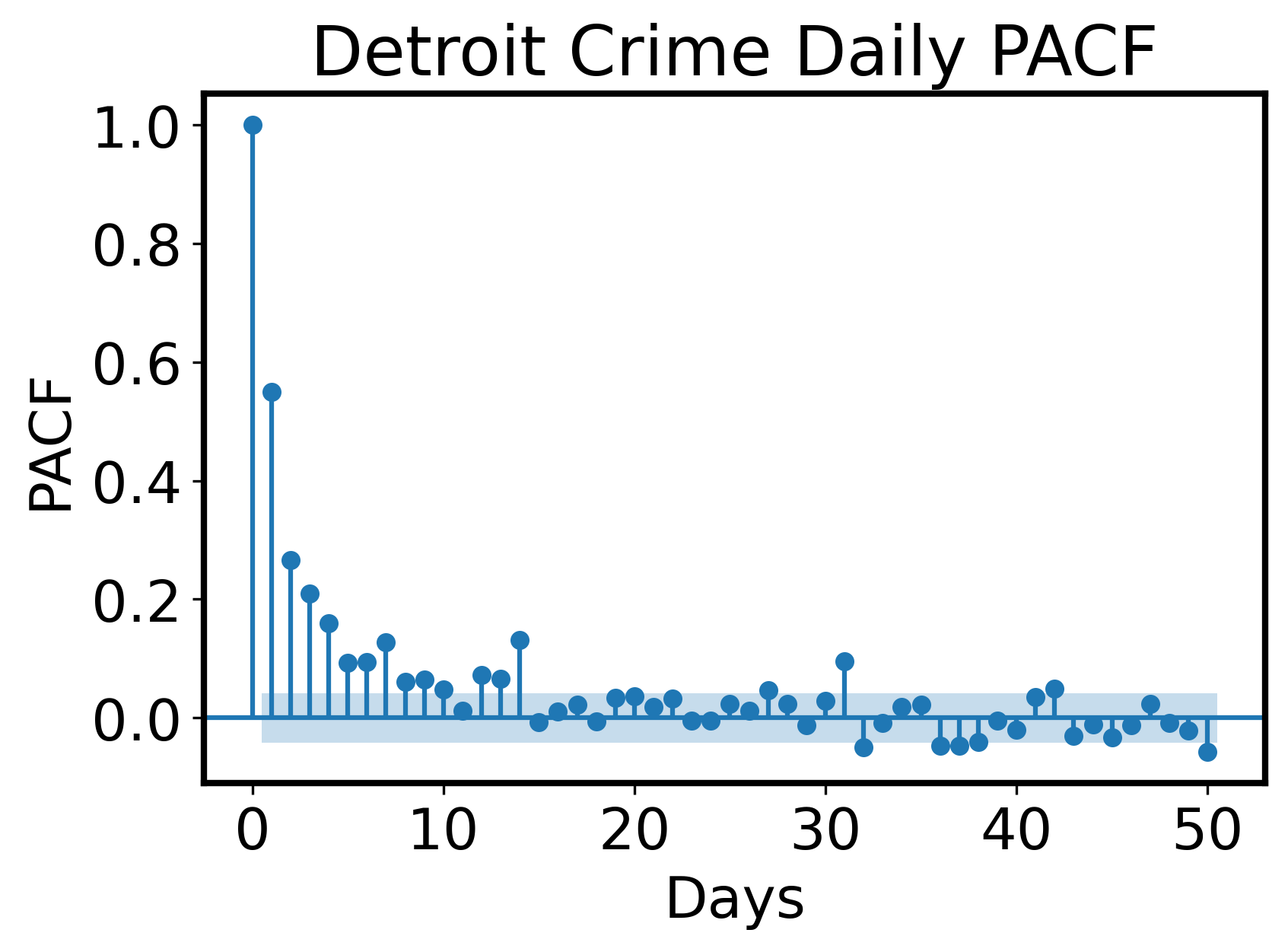}
  \caption{Detroit daily partial autocorrelation (PACF) \\ over a 50 day time period.}
  \label{fig5:7}
\end{subfigure}\hfil 
\captionsetup[subfigure]{oneside,margin={0.5cm,0cm}}
\begin{subfigure}{0.5\linewidth}
  \includegraphics[width=1.0\linewidth]{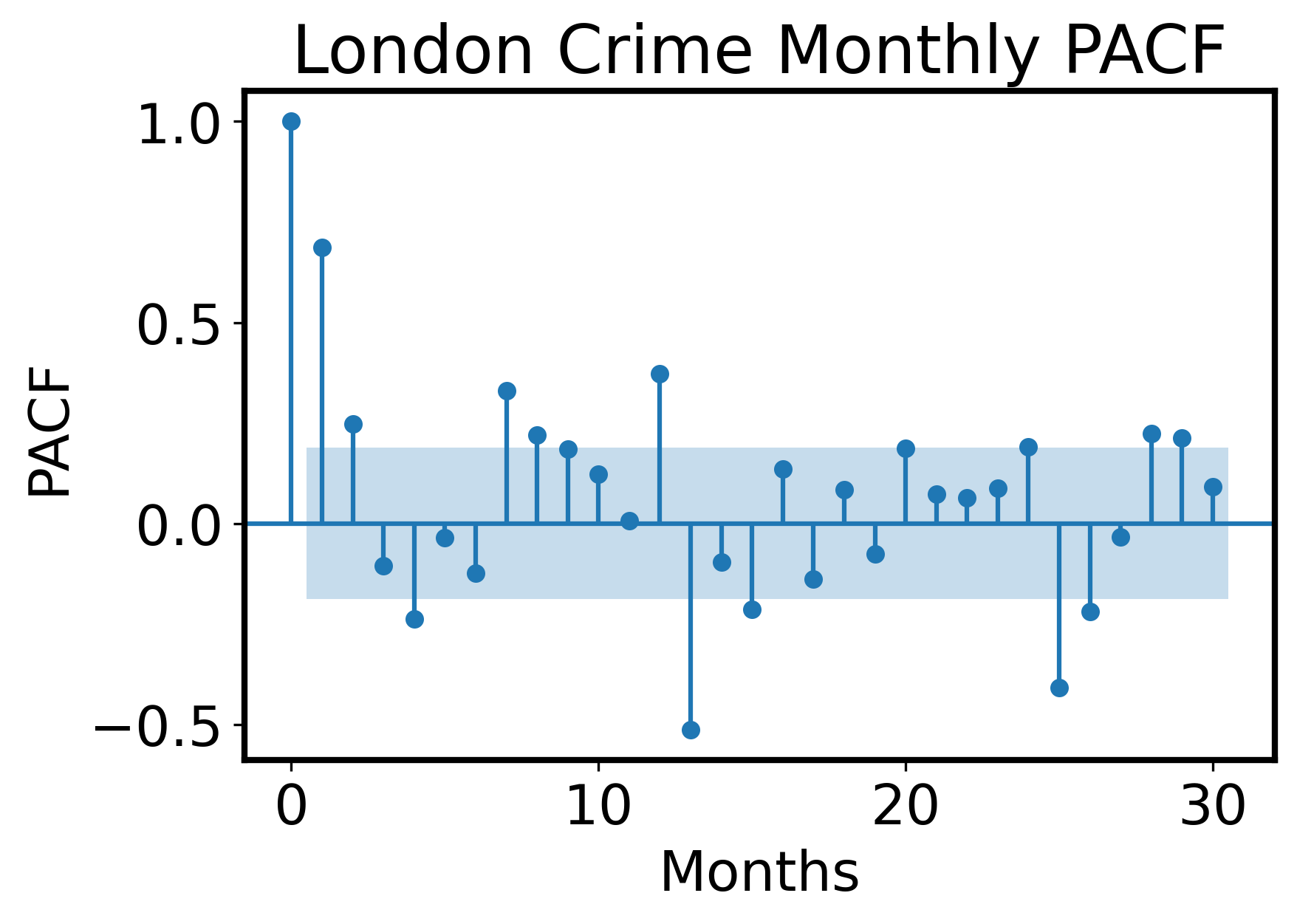}\vspace{1mm}
  \caption{London monthly partial autocorrelation \\ (PACF) over a 30 month time period.}
  \label{fig5:8}
\end{subfigure} 
}
\caption{Partial autocorrelation (PACF) for crime in London (monthly) and Detroit (daily) with shaded blue volume representing a 95\% statistical significance confidence interval.}
\label{pacfthing}
\end{figure}
In \textcolor{blue}{\textbf{Figure \ref{pacfthing}(a)}} it is observed that both the previous and second-most previous observations exhibit statistically significant correlations, indicating that the Detroit dataset is suitable for the application of the history-dependent models. However, in \textcolor{blue}{\textbf{Figure \ref{pacfthing}(b)}} it is evident that only the previous observation shows statistically significant correlation, while the second-most observation does not. This implies that the application of history-dependent models to this dataset is not appropriate.

\subsection{Detroit Crime Model Results}

In a manner akin to Section 5.7, we provide the individual results for each model, following the sequence of their presentation in Section 5.4. Subsequently, we present the consolidated best model results to facilitate a comparative evaluation of the models. This approach allows for a comprehensive assessment of their relative performances.

The outcomes of the vanilla NN model applied to Detroit crime are not included in the analysis as the model exhibited an inability to learn, rendering it capable of predicting only zero for all future estimations, regardless of its underlying structure.

\subsubsection{GWANN Results for Detroit Crime}

The following table displays the achieved evaluation metric scores, including MSE, MAPE, and R2, resulting from the implementation of the GWANN model on the Detroit crime dataset.

\begin{table}[H]
\doublespacing
\centering
\begin{tabular}{cccc}
\hline
\multicolumn{4}{c}{\textbf{GWANN Results for Detroit Crime}} \\ \hline
                & \textbf{MSE} & \textbf{MAPE} & \textbf{R2} \\
1 Hidden Layer  & 1.224        & 1.000         & -0.374      \\
2 Hidden Layers & 1.077        & 0.743         & -0.251      \\
3 Hidden Layers & 0.919        & 0.425         & -0.0785     \\
4 Hidden Layers & 0.919        & 0.421         & -0.0765     \\
5 Hidden Layers & 0.919        & 0.426         & -0.0791    
\end{tabular}
\caption{GWANN results for Detroit crime.}
\end{table}

The results reveal a positive trend in the evaluation metrics as the non-linear complexity of the model increases. This trend shows improvement up to the point of three hidden layers, and from that level onwards, the evaluation metric scores tend to stabilise, remaining nearly constant. The observed significant improvement in MSE scores with increasing non-linear complexity for the Detroit crime dataset, in contrast to the relatively stagnant improvement in MSE scores for the London crime dataset, indicates that the spatial trend for crime in Detroit is more intricate and complex compared to the spatial trend for crime in London.

\subsubsection{GTWNN Results for Detroit Crime}

The following table displays the achieved evaluation metric scores, including MSE, MAPE, and R2, resulting from the implementation of the GTWNN model on the Detroit crime dataset.

\begin{table}[H]
\doublespacing
\centering
\begin{tabular}{cccc}
\hline
\multicolumn{4}{c}{\textbf{GTWNN Results for Detroit Crime}} \\ \hline
                & \textbf{MSE} & \textbf{MAPE} & \textbf{R2} \\
1 Hidden Layer  & 1.089        & 377270912     & 0.0161      \\
2 Hidden Layers & 1.081        & 370377504     & 0.0207      \\
3 Hidden Layers & 1.093        & 372658240     & 0.0140     
\end{tabular}
\caption{GTWNN results for Detroit crime.}
\end{table}

In the obtained results, we observe that GTWNN demonstrates an elevated MSE in comparison to GWANN, and this difference does not show substantial improvement with an increase in the model's non-linear complexity. This observation suggests that the inclusion of external factor information does not consistently result in enhanced predictive capability compared to a spatially expanded output layer. As a consequence, for certain datasets, GWANN may outperform GTWNN in terms of predictive performance. Lastly, it is important to highlight the high MAPE score observed in GTWNN, which exhibits a similar magnitude as the MAPE scores obtained when GTWNN was applied to the London crime dataset. This finding indicates that GTWNN encounters challenges in accurately predicting future instances of low to zero crime occurrences.

\subsubsection{GTWNN\_Ls Results for Detroit Crime}

The following table displays the achieved evaluation metric scores, including MSE, MAPE, and R2, resulting from the implementation of the GTWNN\_Ls model on the Detroit crime dataset.

\begin{table}[H]
\doublespacing
\centering
\begin{tabular}{cccc}
\hline
\multicolumn{4}{c}{\textbf{GTWNN\_Ls Results for Detroit Crime}} \\ \hline
                 & \textbf{MSE}  & \textbf{MAPE}  & \textbf{R2}  \\
1 Hidden Layer   & 7.558         & 0.390          & -0.0865      \\
2 Hidden Layers  & 0.839         & 0.396          & -0.0884      \\
3 Hidden Layers  & 0.841         & 0.401          & -0.0939     
\end{tabular}
\caption{GTWNN\_Ls results for Detroit crime.}
\end{table}

In this context, we observe that increasing the model's non-linear complexity leads to improvements up to a certain threshold (in this case, 2 hidden layers per hidden layer block). When comparing the non-linear complexity of GTWNN\_Ls to GWANN, the predictive capability of GTWNN\_Ls plateaus when the total number of hidden layers reaches 4 (with 2 hidden layers per hidden block). This plateauing effect is quite similar to the results observed in GWANN, where the predictive power stabilises at 3 hidden layers and beyond. However, this plateauing behavior is not evident in the GTWNN\_Ls results obtained from the London crime dataset. This finding further suggests that the spatial trend in Detroit crime exhibits a more intricate and sophisticated pattern compared to London crime.

Moreover, upon comparing GWANN and GTWNN\_Ls, it becomes evident that the inclusion of an intermediate layer does not compromise the predictive power; rather, it generally enhances it. This observation indicates that the incorporation of an intermediate layer in the model contributes positively to its overall performance.

\subsubsection{GTWNN\_Lst Results for Detroit Crime}

The following table displays the achieved evaluation metric scores, including MSE, MAPE, and R2, resulting from the implementation of the GTWNN\_Lst model on the Detroit crime dataset.

\begin{table}[H]
\doublespacing
\centering
\begin{tabular}{cccc}
\hline
\multicolumn{4}{c}{\textbf{GTWNN\_Lst Results for Detroit Crime}} \\ \hline
                  & \textbf{MSE}  & \textbf{MAPE}  & \textbf{R2}  \\
1 Hidden Layer    & 0.912         & 0.401          & -0.0575      \\
2 Hidden Layers   & 0.911         & 0.407          & -0.0613      \\
3 Hidden Layers   & 0.910         & 0.387          & -0.0496     
\end{tabular}
\caption{GTWNN\_Lst results for Detroit crime.}
\end{table}

Upon comparing the Detroit model results for GTWNN\_Ls and GTWNN\_Lst, it becomes apparent that the incorporation of a spatiotemporally expanded output layer does not lead to any significant improvement in predictive power; rather, it slightly diminishes it. This observation suggests the existence of a low temporal correlation between neighboring cells in the Detroit crime dataset.

\subsubsection{HDGTWNN Results for Detroit Crime}

The following table displays the achieved evaluation metric scores, including MSE, MAPE, and R2, resulting from the implementation of the HDGTWNN model on the Detroit crime dataset.

\begin{table}[H]
\doublespacing
\centering
\begin{tabular}{cccc}
\hline
\multicolumn{4}{c}{\textbf{HDGTWNN Results for Detroit Crime}} \\ \hline
                 & \textbf{MSE}  & \textbf{MAPE} & \textbf{R2} \\
1 Hidden Layer   & 1.062         & 370537792     & 0.0349      \\
2 Hidden Layers  & 1.064         & 381253344     & 0.0325      \\
3 Hidden Layers  & 1.069         & 366343392     & 0.0304     
\end{tabular}
\caption{HDGTWNN results for Detroit crime}
\end{table}

Upon comparing these outcomes with GTWNN, it becomes evident that the incorporation of the history dependent module results in improved metric scores across the board without compromising the predictive advantages offered by the model in the absence of the history dependent module. This observation is consistent with the findings observed in the London crime results for GTWNN and HDGTWNN.

\subsubsection{HDGTWNN\_Ls Results for Detroit Crime}

The following table displays the achieved evaluation metric scores, including MSE, MAPE, and R2, resulting from the implementation of the HDGTWNN model on the Detroit crime dataset.

\begin{table}[H]
\doublespacing
\centering
\begin{tabular}{cccc}
\hline
\multicolumn{4}{c}{\textbf{HDGTWNN\_Ls Results for Detroit Crime}} \\ \hline
                  & \textbf{MSE}   & \textbf{MAPE}  & \textbf{R2}  \\
1 Hidden Layer    & 0.912          & 0.396          & -0.0611      \\
2 Hidden Layers   & 0.912          & 0.405          & -0.0646      \\
3 Hidden Layers   & 0.913          & 0.438          & -0.0805     
\end{tabular}
\caption{HDGTWNN\_Ls results for Detroit crime.}
\end{table}

Upon comparing the results of HDGTWNN and HDGTWNN\_Ls, it is evident that the inclusion of a spatially expanded output layer leads to improved evaluation scores overall and enhances the model's ability to handle future instances with low to zero crime more accurately. On the other hand, when comparing HDGTWNN\_Ls with GTWNN\_Ls, the addition of the history dependent module does not result in an increase in predictive power in the model. This observation suggests that, in the context of Detroit crime, a greater emphasis on the spatial correlation of neighboring cells provides more predictive power than a dual focus on both spatial and temporal correlation of neighboring cells.

\subsubsection{HDGTWNN\_Lst Results for Detroit Crime}

The following table displays the achieved evaluation metric scores, including MSE, MAPE, and R2, resulting from the implementation of the HDGTWNN model on the Detroit crime dataset.

\begin{table}[H]
\doublespacing
\centering
\begin{tabular}{cccc}
\hline
\multicolumn{4}{c}{\textbf{HDGTWNN\_Lst Results for Detroit Crime}} \\ \hline
                  & \textbf{MSE}   & \textbf{MAPE}   & \textbf{R2}  \\
1 Hidden Layer    & 0.912          & 0.392           & -0.0463      \\
2 Hidden Layers   & 0.911          & 0.389           & -0.0436      \\
3 Hidden Layers   & 0.910          & 0.401           & -0.0490     
\end{tabular}
\caption{HDGTWNN\_Lst results for Detroit crime.}
\end{table}

The evaluation metric scores obtained for Detroit crime using HDGTWNN\_Lst are nearly identical to the scores obtained with HDGTWNN\_Ls. This indicates that employing both a history dependent module and a spatiotemporally expanded output layer to leverage temporal correlation between neighboring cells as a means to enhance predictive power may be redundant. This observation aligns with the results observed in the analysis of London crime data for HDGTWNN\_Ls and HDGTWNN\_Lst.

\subsubsection{Aggregated Best Model Results for Detroit crime}

Here, we present the aggregated results for all the ANN models applied to the Detroit crime dataset. Each entry in the table represents the lowest MSE score obtained from our NAS strategy, accompanied by their corresponding MAPE and R2 scores. The models are arranged in descending order based on their MSE values.

\begin{table}[H]
\doublespacing
\centering
\begin{tabular}{cccc}
\hline
\multicolumn{4}{c}{\textbf{Best Model Results for Detroit Crime}} \\ \hline
\textbf{}      & \textbf{MSE}   & \textbf{MAPE}   & \textbf{R2}   \\
GTWNN          & 1.081          & 370377504       & 0.0207        \\
HDGTWNN        & 1.062          & 370537792       & 0.0349        \\
GWANN          & 0.919          & 0.421           & -0.0765       \\
HDGTWNN\_Ls    & 0.912          & 0.396           & -0.0611       \\
HDGTWNN\_Lst   & 0.910          & 0.401           & -0.0490       \\
GTWNN\_Lst     & 0.910          & 0.387           & -0.0496       \\
GTWNN\_Ls      & 0.839          & 0.396           & -0.0884       
\end{tabular}
\caption{Aggregated best model results for Detroit crime arranged in order of decreasing MSE scores.}
\end{table}

Among the models, the GTWNN\_Ls achieves the smallest MSE score, while the GTWNN obtains the largest MSE score. This indicates that for this dataset, while a non-linear combination of external factor information (via an intermediate input layer) affords the model some generally increased degree of predictive power, a strong emphasis on spatial correlation between neighboring cells, is crucial. In essence, these findings suggest that the Detroit crime dataset exhibits low temporal correlation and high spatial correlation between neighboring cells. 

\subsection{Spatial Autocorrelation of London and Detroit Crime}

In this subsection, we present two key conclusions drawn from the model results:

\begin{enumerate}
    \item London crime exhibits low spatial and high temporal correlation among neighboring cells.
    \item Detroit crime, on the other hand, demonstrates high spatial and low temporal correlation among neighboring cells.
\end{enumerate}

In section 5.10 we presented temporal PACF plots. The nearest neighbor temporal correlation results for Detroit and London were found to be 0.55 and 0.69, respectively, which aligns with the aforementioned conclusions. To test if the spatial correlations also fit the aforementioned conclusion spatial PACF plots were generated along both the horizontal and vertical axes for both datasets,

\begin{figure}[H]
  \captionsetup[subfigure]{margin={1cm,1cm},justification=justified,singlelinecheck=false}
\makebox[\linewidth]{
\begin{subfigure}{0.5\textwidth}
  \centering
  \includegraphics[width=1.0\linewidth]{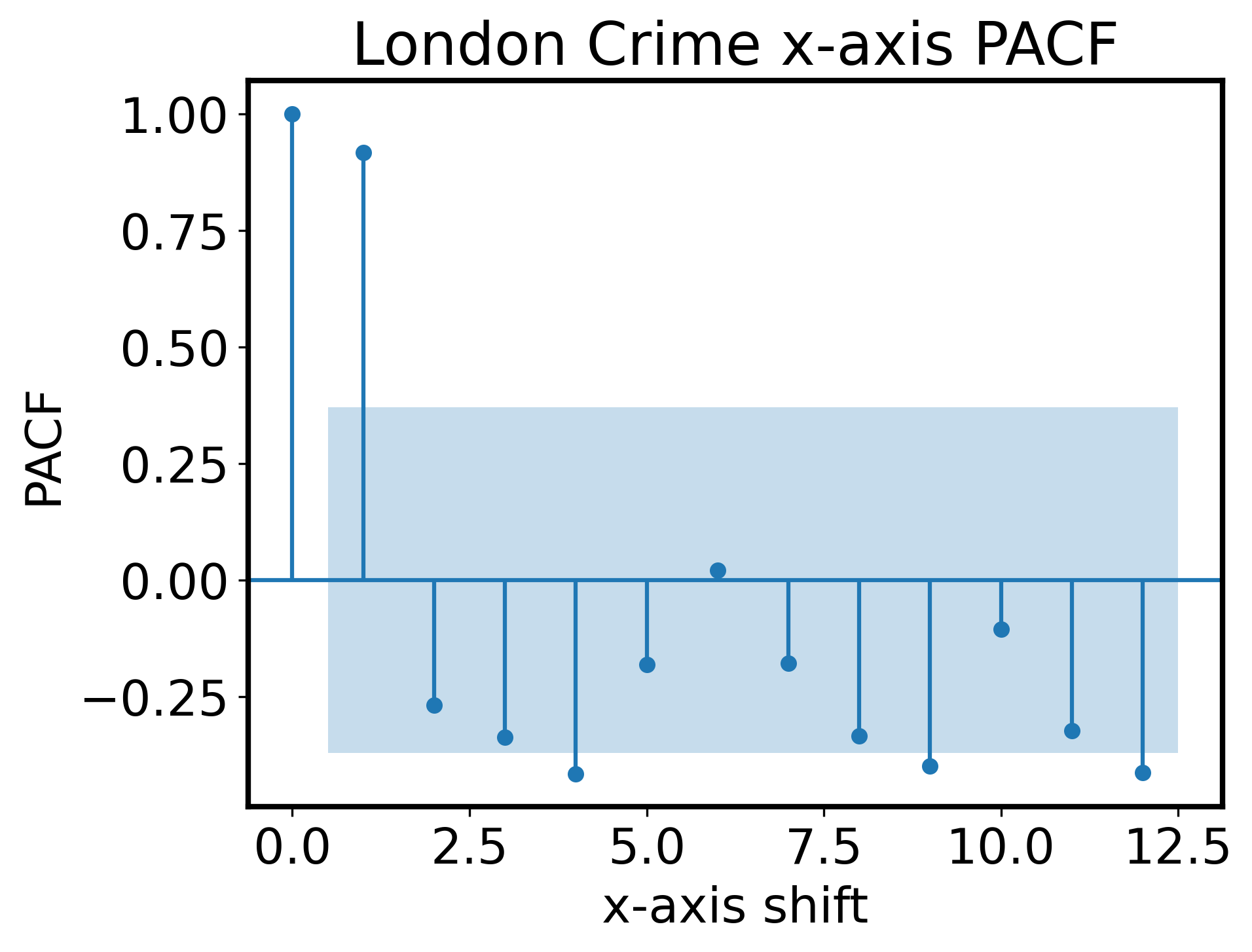} 
  \caption{Spatial PACF along the horizontal x-axis for London crime.}
  \label{fig5:9}
\end{subfigure}\hfil 
\begin{subfigure}{0.5\textwidth}
  \centering
  \includegraphics[width=1.0\linewidth]{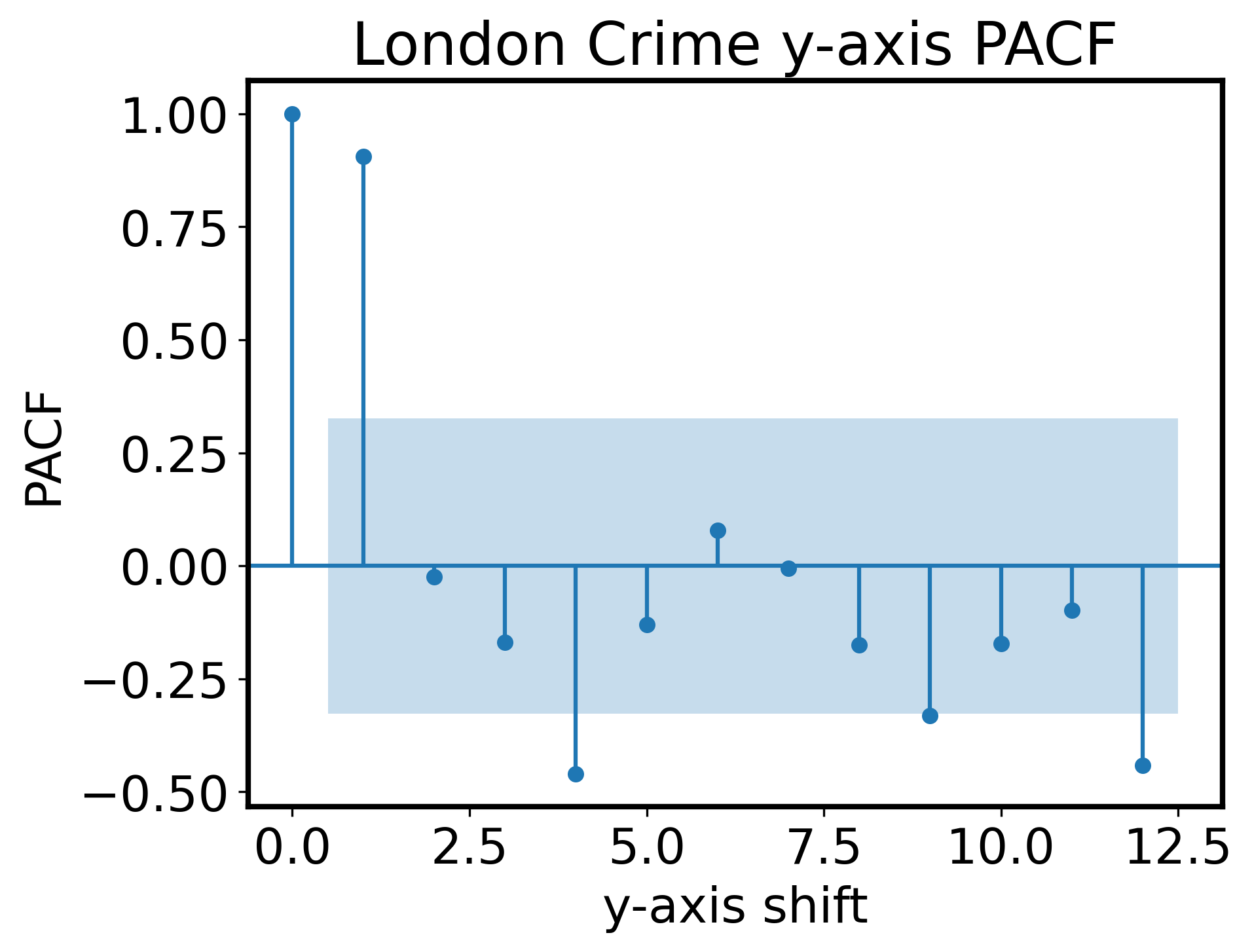}
  \caption{Spatial PACF along the vertical y-axis for London crime.}
  \label{fig5:10}
\end{subfigure} 
}
\makebox[\linewidth]{
\begin{subfigure}{0.5\textwidth}
  \centering
  \includegraphics[width=1.0\linewidth]{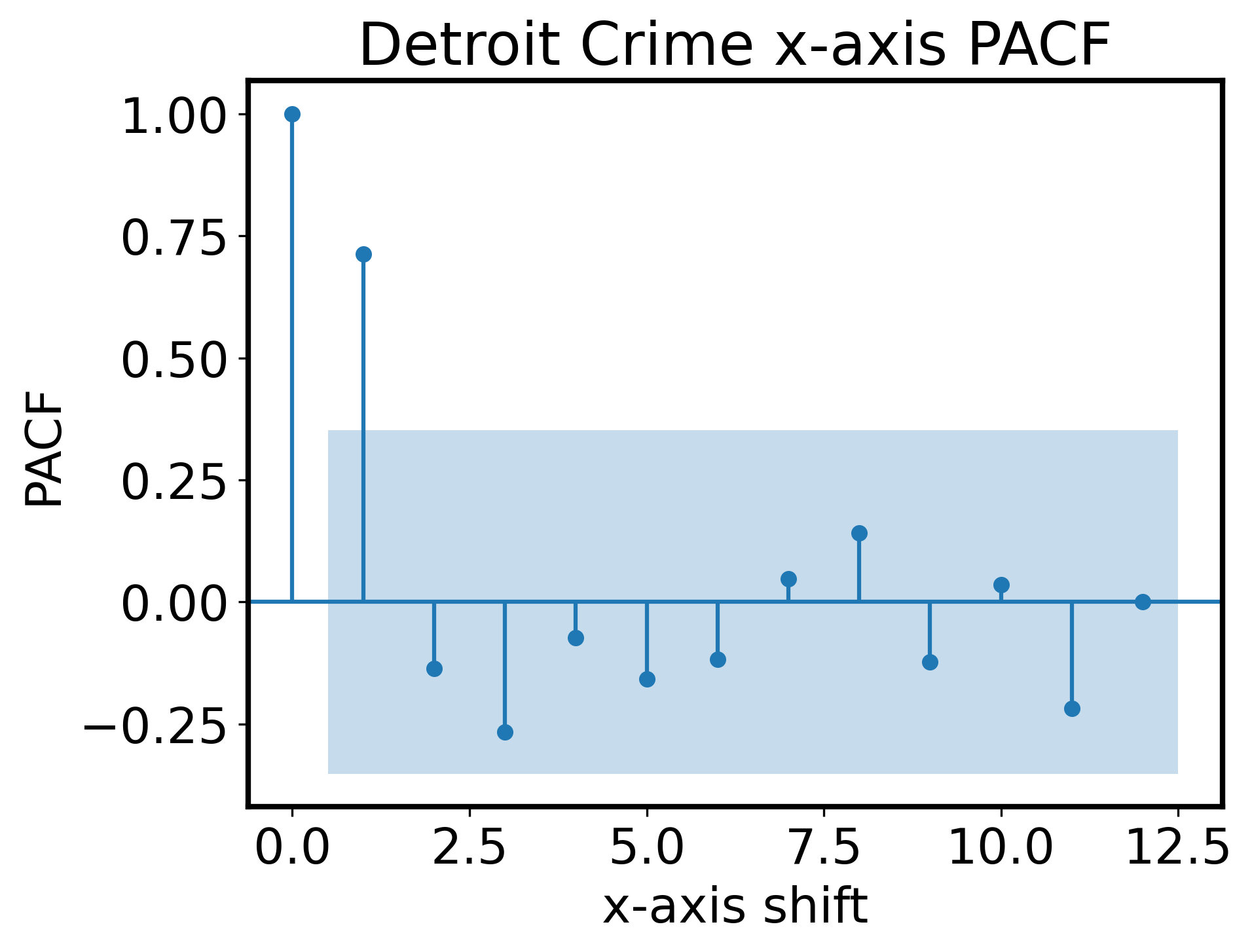} 
  \caption{Spatial PACF along the horizontal x-axis for Detroit crime.}
  \label{fig5:11}
\end{subfigure}\hfil 
\begin{subfigure}{0.5\textwidth}
  \centering
  \includegraphics[width=1.0\linewidth]{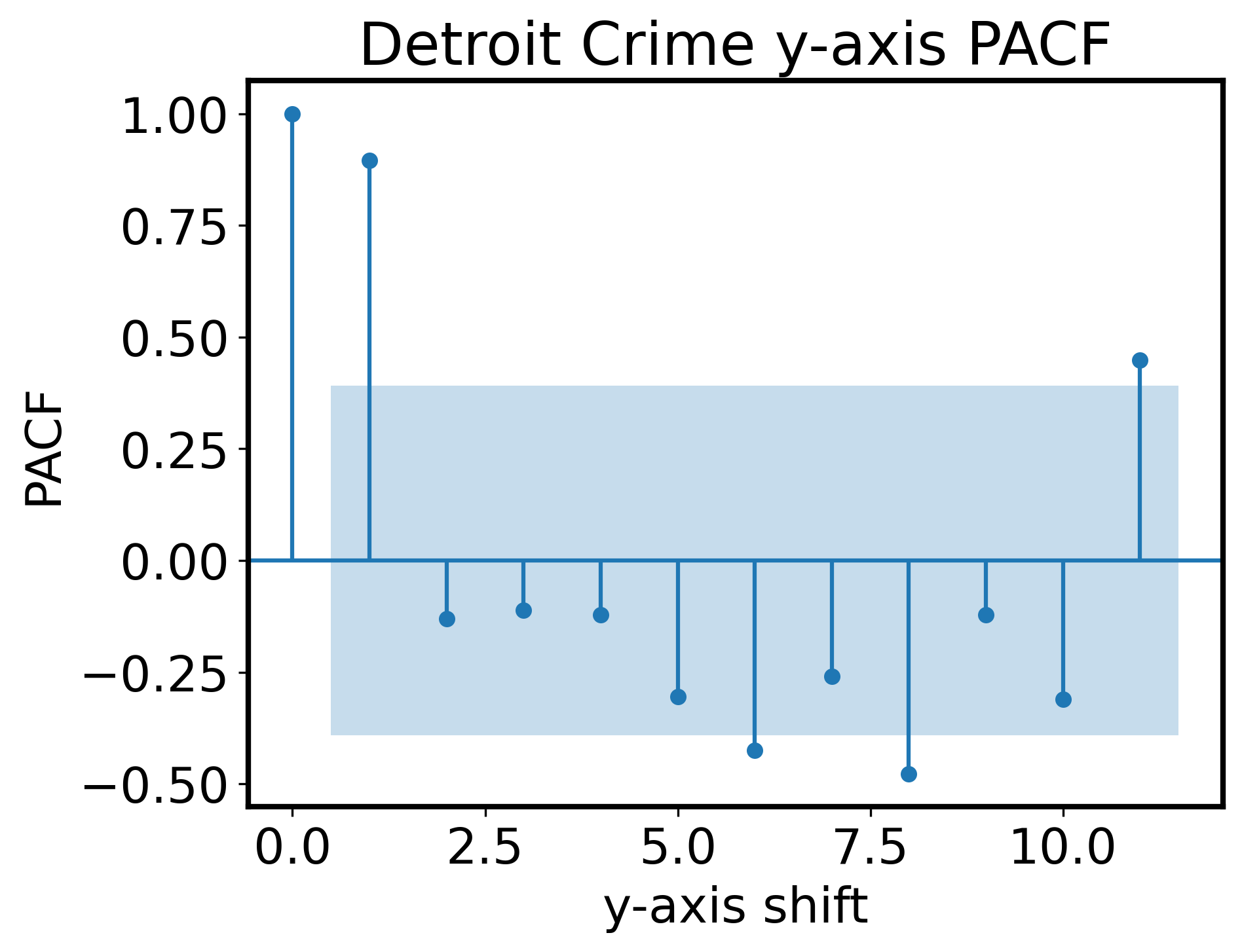}
  \caption{Spatial PACF along the vertical y-axis for Detroit crime.}
  \label{fig5:12}
\end{subfigure}
}
\caption{PACF plots for both London and Detroit along both spatial axes.}
\label{fig5-13}
\end{figure}

revealing neighboring cell spatial correlations as follows:

\begin{itemize}
    \item London Neighbouring cell spatial x-axis PACF =  0.92
    \item London Neighbouring cell spatial y-axis PACF = 0.90
    \item Detroit Neighbouring cell spatial x-axis PACF = 0.71
    \item Detroit Neighbouring cell spatial y-axis PACF = 0.89
\end{itemize}

These findings may initially appear to deviate from the initial narrative presented at the beginning of this subsection. However, it is important to approach these results with a degree of critical evaluation. To better understand the origins of the conclusions regarding low and high spatial correlations in London and Detroit, we must consider the methodology employed. The inference of spatial correlation was primarily based on the MSE scores obtained from the models applied to both datasets.

It is essential to recognise that a very high MAPE can indicate poor predictive capabilities for instances with low to zero future crimes, while a very high MSE signifies poor predictive capabilities for instances with a high count of future crimes. Therefore, the conclusions drawn about neighboring cell spatial correlations based on MSE values primarily apply to instances with high crime counts.

As a result, we have refined the two initial conclusions from the results section to provide a more precise understanding of the nature of spatial correlations present in the datasets:

\begin{enumerate}
    \item London crime has low spatial neighbouring cell correlation around high density crime locations.
    \item Detroit crime has high spatial neighbouring cell correlation around high density crime locations.
\end{enumerate}

\begin{figure}[H]
  \captionsetup[subfigure]{oneside,margin={1cm,1cm}}
\makebox[\linewidth]{
\begin{subfigure}{0.5\textwidth}
  \centering
  \includegraphics[width=1.0\linewidth]{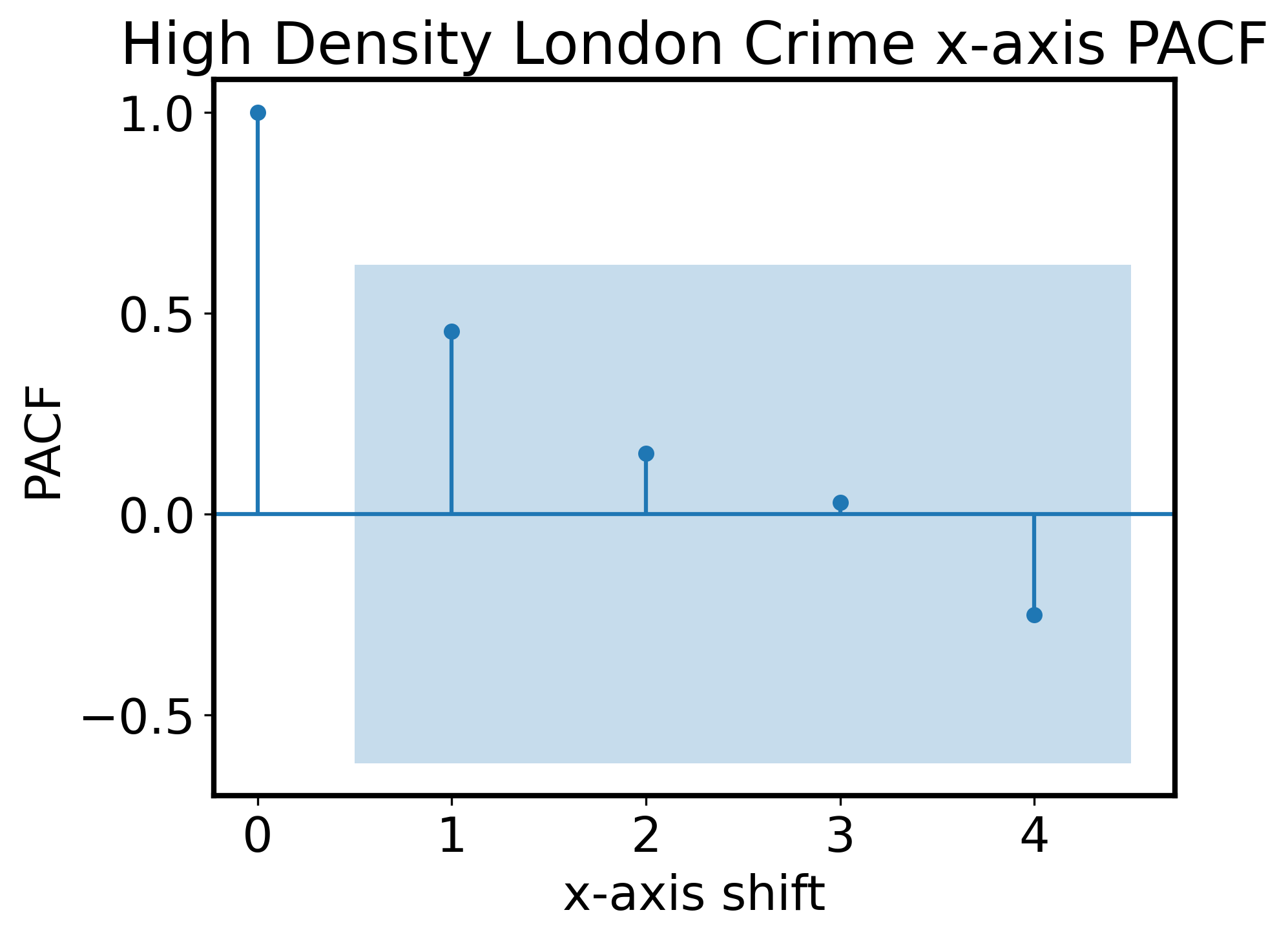} 
  \caption{Spatial PACF along the horizontal x-axis for high density London crime locations.}
  \label{fig5:14}
\end{subfigure}\hfil 
\begin{subfigure}{0.5\textwidth}
  \centering
  \includegraphics[width=1.0\linewidth]{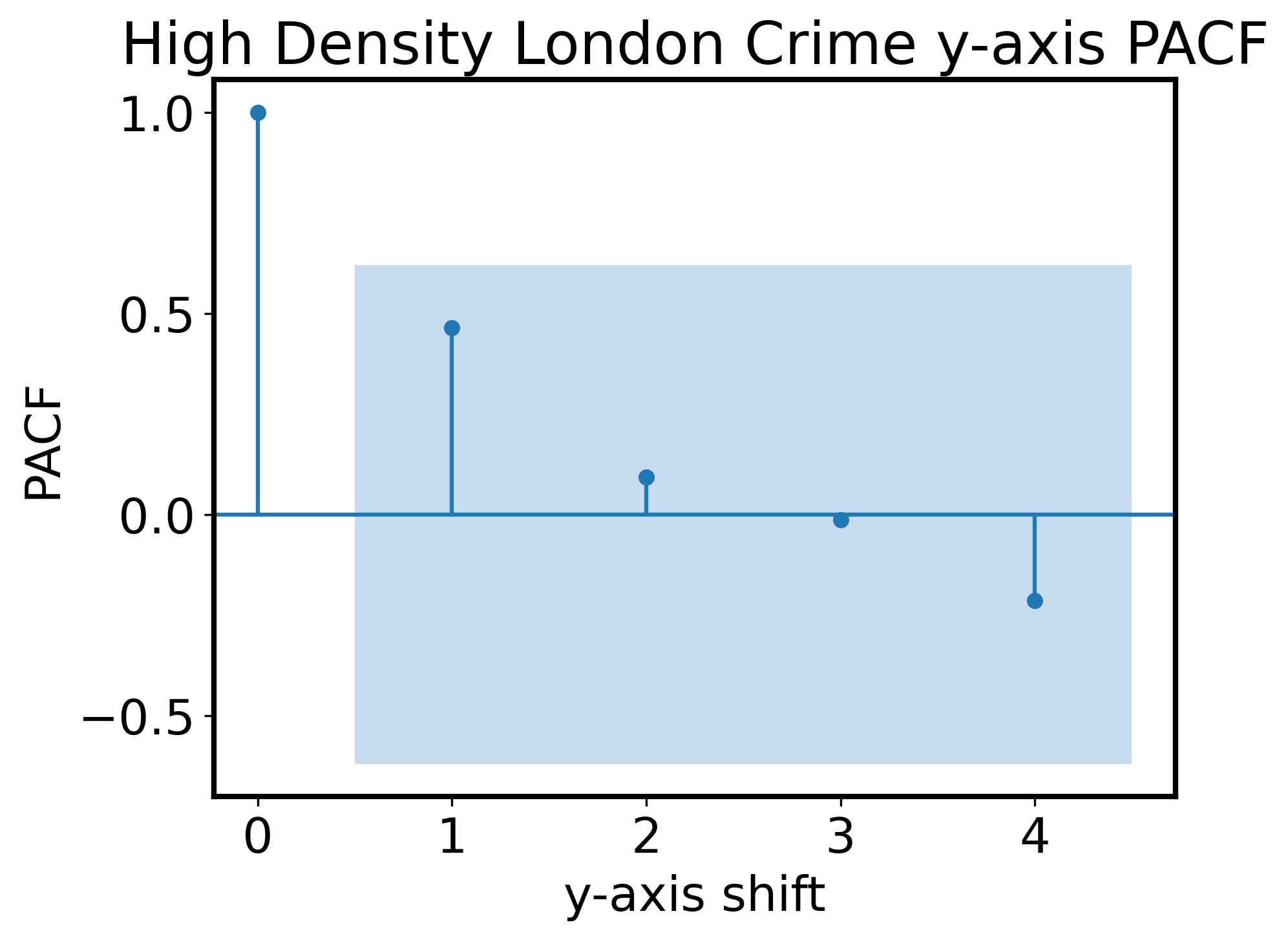}
  \caption{Spatial PACF along the vertical y-axis for high density London crime locations.}
  \label{fig5:15}
\end{subfigure} 
}
\makebox[\linewidth]{
\begin{subfigure}{0.5\textwidth}
  \centering
  \includegraphics[width=1.0\linewidth]{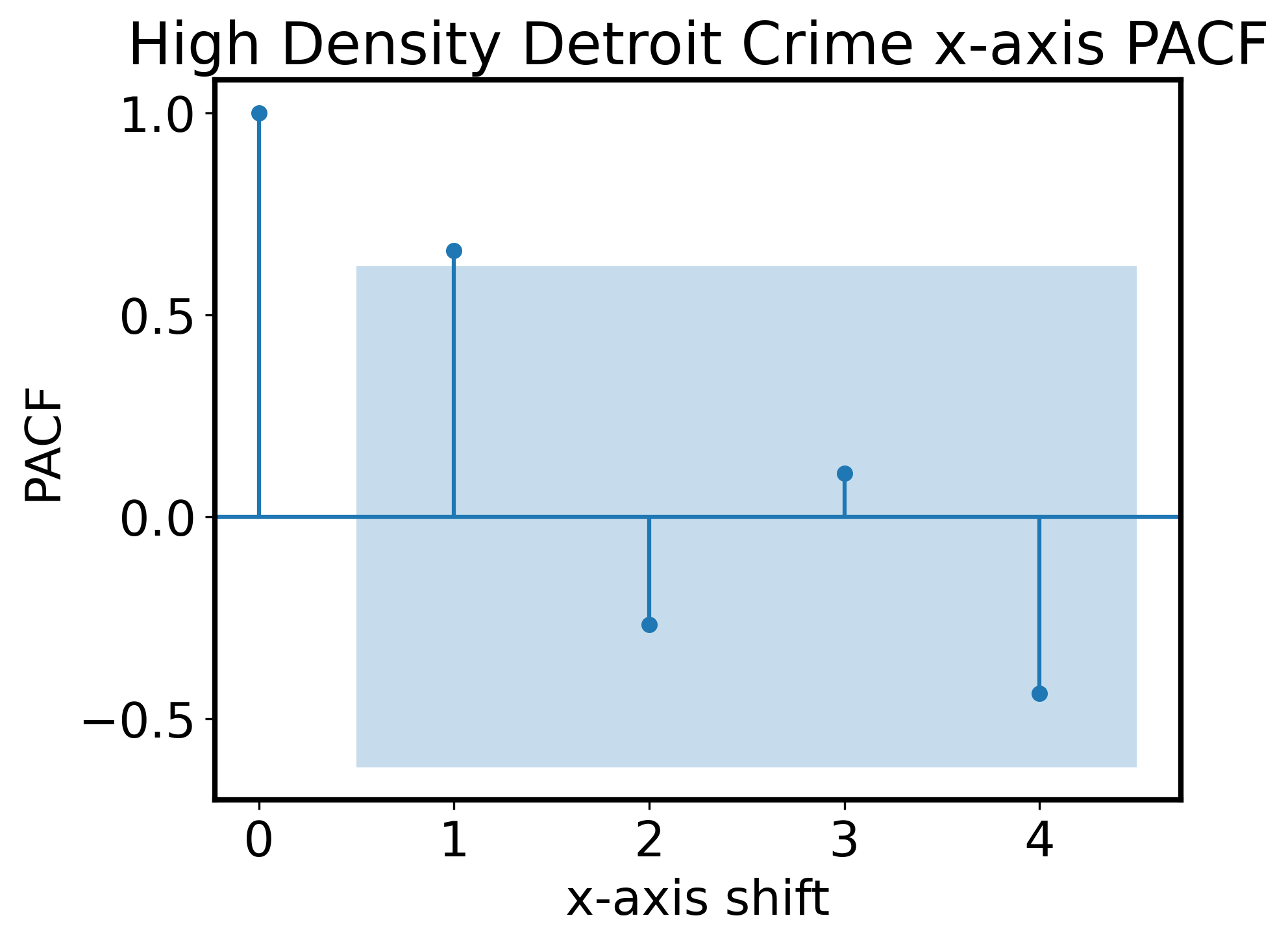} 
  \caption{Spatial PACF along the horizontal x-axis for high density Detroit crime locations.}
  \label{fig6:1}
\end{subfigure}\hfil 
\begin{subfigure}{0.5\textwidth}
  \centering
  \includegraphics[width=1.0\linewidth]{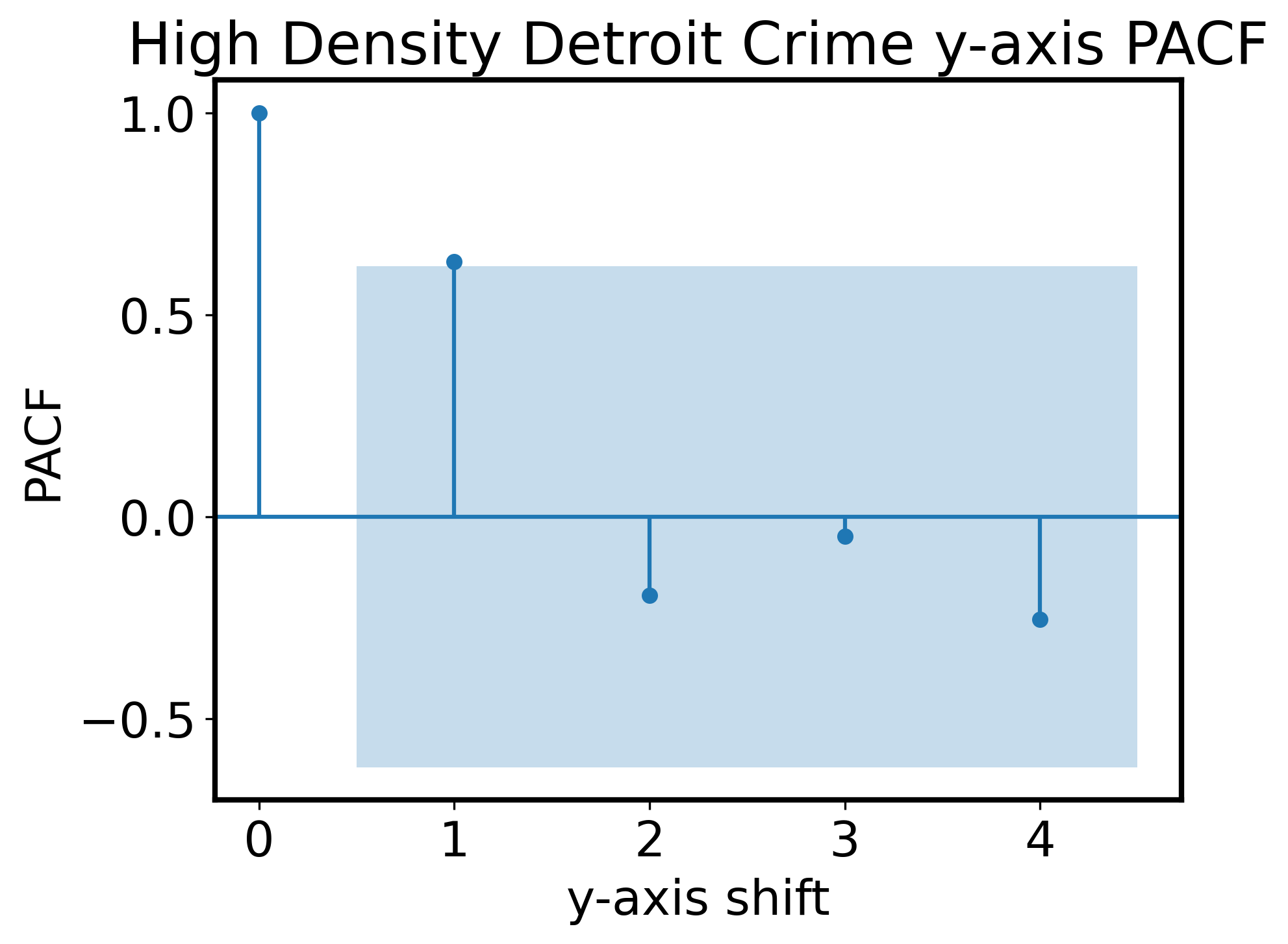}
  \caption{Spatial PACF along the vertical y-axis for high density Detroit crime locations.}
  \label{fig6:2}
\end{subfigure}
}
\caption{PACF plots for high density locations in London and Detroit along both spatial axes.}
\label{fig6}
\end{figure}

To verify these revised conclusions, we conducted an investigation by isolating the top 10 horizontal and vertical slices that encompass the highest number of total crimes in both datasets. Our aim was to examine the nearest neighbor PACF and assess its statistical significance for both Detroit and London. We anticipated that if our amended conclusion is accurate, we would observe a statistically significant nearest neighbor PACF for Detroit, but not for London. The spatial PACF plots along both spatial axes, specifically for high-density crime locations in London and Detroit, are presented in Figure \ref{fig6:2}.

From which, we report the values of the neighboring cell spatial correlation for high-density crime locations in London and Detroit along both spatial axes:

\begin{itemize}
    \item High density London crime neighbouring cell spatial x-axis PACF =  0.45
    \item High density London crime neighbouring cell spatial y-axis PACF = 0.46
    \item High density Detroit crime neighbouring cell spatial x-axis PACF = 0.77
    \item High density Detroit crime Neighbouring cell spatial y-axis PACF = 0.63
\end{itemize}

These results align with the revised explanation. Considering that the MSE score is particularly sensitive to predictions in high-density crime locations, and given the statistically significant neighboring cell spatial PACF observed for high-density crime locations in Detroit, models that incorporate spatial correlation are anticipated to yield lower MSE scores. This observation is in line with the outcomes reported in section 5.10.

On the contrary, the spatial PACFs for neighboring cells in high-density London crime locations are not statistically significant. Consequently, the incorporation of elements that consider spatial correlation may enhance the model's performance for low to zero future crime instances (given the overall high and statistically significant spatial PACF for neighbouring cells of general London crime locations). However, since neighbouring cell spatial correlation lacks statistical significance for high-density crime locations, the inclusion of elements accounting for spatial correlation is not likely to result in a significant reduction in MSE. This observation is consistent with the findings reported in section 5.9.

\section{Conclusions}

It is essential to highlight that the models presented in this section, although applied specifically to crime data from London and Detroit, demonstrate concepts that hold broader applicability to spatiotemporal prediction problems. The methodologies and insights discussed herein can be extended and adapted to address prediction challenges in diverse geographic and temporal contexts.

In Section 5.3, we introduced and delineated three novel enhancements to the current state-of-the-art model, GTWNN. The results presented in sections 5.7 and 5.9 indicated that, among these enhancements, the combination of extensions 2 (spatiotemporally expanded output layer and spatiotemporally weight loss function) and 3 (the history dependent module) proved to be redundant. Furthermore, it was observed that when considering the incorporation of neighbouring cell temporal correlation information, the history dependent module performed better in integrating the relevant temporal correlations without compromising the other advantageous features that the model presented in the absence of the history dependent module.

Furthermore, we demonstrated that the consideration of neighbouring cell temporal and spatial correlations plays a crucial role in determining the suitable type of extended framework for a given spatiotemporal problem. When neighbouring cells exhibit significant temporal correlation, the inclusion of a history dependent module proves to be appropriate. On the other hand, if neighbouring cells display notable spatial correlation, the adoption of a spatially expanded output layer coupled with a spatially weighted loss function (extension 1, section 5.3.1) may be warranted. Moreover, in scenarios where the neighbouring cell PACF is large for general locations, the network gains enhanced predictive capabilities in handling future predictions of low to zero crime instances. Conversely, if neighbouring cells surrounding high-density crime locations exhibit a high PACF, incorporating extension 1 is likely to result in lower MSE scores, as evidenced in section 5.10.

Consequently, we put forth the following networks as potential state-of-the-art models for spatiotemporal problems:

\begin{enumerate}
    \item GTWNN\_Ls: Recommended for spatiotemporal prediction tasks characterised by high values of neighbouring cell spatial PACF.
    \item HDGTWNN: Suited for spatiotemporal prediction problems with notable neighbouring cell temporal PACF values.
    \item HDGTWNN\_Ls: Appropriate for spatiotemporal prediction problems exhibiting both high neighbouring cell temporal and spatial PACF values.
\end{enumerate}

\bibliography{crime}

\begin{thebibliography}{}

\bibitem[Bernasco and Nieuwbeerta, 2005]{bernasco2005residential}
Bernasco, W. and Nieuwbeerta, P. (2005).
\newblock How do residential burglars select target areas? a new approach to the analysis of criminal location choice.
\newblock {\em British Journal of Criminology}, 45(3):296--315.

\bibitem[Brantingham and Brantingham, 2013]{brantingham2013crime}
Brantingham, P. and Brantingham, P. (2013).
\newblock Crime pattern theory.
\newblock In {\em Environmental criminology and crime analysis}, pages 100--116. Willan.

\bibitem[Brunsdon et~al., 1998]{brunsdon1998geographically}
Brunsdon, C., Fotheringham, S., and Charlton, M. (1998).
\newblock Geographically weighted regression.
\newblock {\em Journal of the Royal Statistical Society: Series D (The Statistician)}, 47(3):431--443.

\bibitem[Cohen and Felson, 1979]{cohen1979social}
Cohen, L.~E. and Felson, M. (1979).
\newblock Social change and crime rate trends: A routine activity approach.
\newblock {\em American sociological review}, pages 588--608.

\bibitem[Cornish and Clarke, 2014]{cornish2014reasoning}
Cornish, D.~B. and Clarke, R.~V. (2014).
\newblock {\em The reasoning criminal: Rational choice perspectives on offending}.
\newblock Transaction Publishers.

\bibitem[Feng et~al., 2021]{feng2021geographically}
Feng, L., Wang, Y., Zhang, Z., and Du, Q. (2021).
\newblock Geographically and temporally weighted neural network for winter wheat yield prediction.
\newblock {\em Remote Sensing of Environment}, 262:112514.

\bibitem[Ferreira et~al., 2012]{ferreira2012gis}
Ferreira, J., Jo{\~a}o, P., and Martins, J. (2012).
\newblock Gis for crime analysis: Geography for predictive models.
\newblock {\em Electronic Journal of Information Systems Evaluation}, 15(1):pp36--49.

\bibitem[Fotheringham et~al., 2015]{fotheringham2015geographical}
Fotheringham, A.~S., Crespo, R., and Yao, J. (2015).
\newblock Geographical and temporal weighted regression (gtwr).
\newblock {\em Geographical Analysis}, 47(4):431--452.

\bibitem[Frank et~al., 2012]{frank2012criminal}
Frank, R., Andresen, M.~A., and Brantingham, P.~L. (2012).
\newblock Criminal directionality and the structure of urban form.
\newblock {\em Journal of Environmental Psychology}, 32(1):37--42.

\bibitem[Fyfe, 2000]{fyfe2000geography}
Fyfe, N. (2000).
\newblock Geography of crime.
\newblock {\em The dictionary of human geography. London (Blackwell Publishers)}, page 121.

\bibitem[Fyfe and Reeves, 2011]{fyfe2011thin}
Fyfe, N.~R. and Reeves, A.~D. (2011).
\newblock The thin green line? police perceptions of the challenges of policing wildlife crime in scotland.
\newblock In {\em Rural policing and policing the rural: a constable countryside?}, pages 169--182. Ashgate.

\bibitem[Golledge et~al., 1987]{golledge1987analytical}
Golledge, R.~G., Golledge, R., and Stimson, R.~J. (1987).
\newblock {\em Analytical behavioural geography}.
\newblock Routledge Kegan \& Paul.

\bibitem[Hagenauer and Helbich, 2022]{hagenauer2022geographically}
Hagenauer, J. and Helbich, M. (2022).
\newblock A geographically weighted artificial neural network.
\newblock {\em International Journal of Geographical Information Science}, 36(2):215--235.

\bibitem[Jenga et~al., 2023]{jenga2023machine}
Jenga, K., Catal, C., and Kar, G. (2023).
\newblock Machine learning in crime prediction.
\newblock {\em Journal of Ambient Intelligence and Humanized Computing}, 14(3):2887--2913.

\bibitem[Kingma and Ba, 2014]{kingma2014adam}
Kingma, D.~P. and Ba, J. (2014).
\newblock Adam: A method for stochastic optimization.
\newblock {\em arXiv preprint arXiv:1412.6980}.

\bibitem[Radford et~al., 2018]{radford2018improving}
Radford, A., Narasimhan, K., Salimans, T., Sutskever, I., et~al. (2018).
\newblock Improving language understanding by generative pre-training.
\newblock {\em OpenAI}.

\bibitem[Ramesh et~al., 2022]{ramesh2022hierarchical}
Ramesh, A., Dhariwal, P., Nichol, A., Chu, C., and Chen, M. (2022).
\newblock Hierarchical text-conditional image generation with clip latents.
\newblock {\em arXiv preprint arXiv:2204.06125}, 1(2):3.

\bibitem[Reid et~al., 2014]{reid2014uncovering}
Reid, A.~A., Frank, R., Iwanski, N., Dabbaghian, V., and Brantingham, P. (2014).
\newblock Uncovering the spatial patterning of crimes: A criminal movement model (crimm).
\newblock {\em Journal of research in crime and delinquency}, 51(2):230--255.

\bibitem[Rummens et~al., 2017]{rummens2017use}
Rummens, A., Hardyns, W., and Pauwels, L. (2017).
\newblock The use of predictive analysis in spatiotemporal crime forecasting: Building and testing a model in an urban context.
\newblock {\em Applied geography}, 86:255--261.

\bibitem[Sherman and Eck, 2003]{sherman2003policing}
Sherman, L.~W. and Eck, J.~E. (2003).
\newblock Policing for crime prevention.
\newblock In {\em Evidence-based crime prevention}, pages 309--343. Routledge.

\bibitem[Tamir et~al., 2021]{tamir2021crime}
Tamir, A., Watson, E., Willett, B., Hasan, Q., and Yuan, J.-S. (2021).
\newblock Crime prediction and forecasting using machine learning algorithms.
\newblock {\em International Journal of Computer Science and Information Technologies}, 12(2):26--33.

\bibitem[Valentine, 1989]{valentine1989geography}
Valentine, G. (1989).
\newblock The geography of women's fear.
\newblock {\em Area}, pages 385--390.

\bibitem[Wu et~al., 2021]{wu2021geographically}
Wu, S., Wang, Z., Du, Z., Huang, B., Zhang, F., and Liu, R. (2021).
\newblock Geographically and temporally neural network weighted regression for modeling spatiotemporal non-stationary relationships.
\newblock {\em International Journal of Geographical Information Science}, 35(3):582--608.

\bibitem[Yarwood, 2015]{yarwood2015geography}
Yarwood, R. (2015).
\newblock Geography of crime.
\newblock In {\em Oxford Bibliographies in Geography}. Oxford University Press.

\bibitem[Zhang et~al., 2022]{zhang2022interpretable}
Zhang, X., Liu, L., Lan, M., Song, G., Xiao, L., and Chen, J. (2022).
\newblock Interpretable machine learning models for crime prediction.
\newblock {\em Computers, Environment and Urban Systems}, 94:101789.

\end{thebibliography}
\bibliographystyle{apalike}
\end{singlespace}
\end{document}